\documentclass{article}

\usepackage{arxiv}

\usepackage[utf8]{inputenc} 
\usepackage[T1]{fontenc}    
\usepackage{hyperref}       
\usepackage{url}            
\usepackage{booktabs}       
\usepackage{amsfonts}       
\usepackage{nicefrac}       
\usepackage{microtype}      
\usepackage{lipsum}
\usepackage{graphicx}
\graphicspath{ {./images/} }
\usepackage[utf8]{inputenc} 
\usepackage[T1]{fontenc}    
\usepackage{hyperref}       
\usepackage{url}            
\usepackage{booktabs}       
\usepackage{amsfonts}       
\usepackage{nicefrac}       
\usepackage{microtype}      
\usepackage{xcolor}         
\usepackage{float}

\usepackage[round]{natbib}

\usepackage{amsmath, amssymb, xcolor, amsthm}
\usepackage{bbold}
\usepackage{comment}
\usepackage{algorithm}
\usepackage{algorithmic}
\usepackage{subcaption}
\usepackage{multicol}
\def \w {\widehat}
\def \bs {\boldsymbol}
\def \bit{\begin{itemize}}
\def \eit{\end{itemize}}
\def \ind{\mathbb 1}
\def \m{\mathbf}

\usepackage[utf8]{inputenc}
\usepackage{graphicx}
\usepackage{array}

\title{Generalized Oversampling for Learning from Imbalanced datasets and Associated Theory}

\author{
  Samuel Stocksieker  \\
  Univ. Lyon, UCBL, ISFA LSAF \\
   \And
 Denys Pommeret  \\
  Aix-Marseille Univ., CNRS, I2M \\
  \And
 Arthur Charpentier \\
  UQAM - Montréal \\
}

\begin{document}
\maketitle

\begin{abstract}
In supervised learning, it is quite frequent to be confronted with real imbalanced datasets. This situation leads to a learning difficulty for standard algorithms. Research and solutions in imbalanced learning have mainly focused on classification tasks. Despite its importance, very few solutions exist for imbalanced regression. 
In this paper, we propose a data augmentation procedure, the GOLIATH algorithm, based on kernel density estimates which can be used in classification and regression. 
This general approach encompasses  two large families of synthetic oversampling: those based on perturbations, such as  Gaussian Noise, and those based on interpolations, such as SMOTE. It also provides an explicit form of these machine learning algorithms and an expression of their conditional densities, in particular for SMOTE. New synthetic data generators  are deduced. We apply GOLIATH in imbalanced regression  combining such generator procedures with a wild-bootstrap resampling technique for the target values. We  evaluate the performance of the GOLIATH algorithm in  imbalanced regression situations. 
We empirically evaluate and compare our approach and demonstrate significant improvement over existing state-of-the-art techniques.

\end{abstract}

\section{Introduction}

Many real-world forecasting problems are based on predictive models in a supervised learning framework and the standard algorithms fail when the target variable is skewed. The learning from imbalanced data concerns many problems with numerous applications in different fields \cite{krawczyk2016learning}, \cite{fernandez2018learning}.
The major part of such works concerns imbalanced classification (see for instance \cite{buda2018} or \cite{cao2019} or \cite{cui2019} or \cite{huang2016} or \cite{yang2020}. As shown by \cite{Branco2016Survey}, many solutions for dealing with the imbalanced learning propose a pre-processing strategy especially the generation of the new synthetic data. A large part of these existing methods consist in combining the well know SMOTE algorithm \cite{fernandez2018smote}.


However, regression tasks over imbalanced data are not as well explored. Few works have addressed the problem despite the importance of this topic. The first and main works on this topic propose to binarize the problem with a relevant function and an associate threshold \cite{torgo2007utility} in order to adapt some Imbalanced classification solutions \cite{torgo2013smote}, \cite{branco2017smogn},  \cite{branco2019pre}, \cite{ribeiro2020imbalanced}, \cite{song2022distsmogn}, \cite{camacho2022geometric} for instance. This methodology presents the disadvantage of dividing the continuous distribution of the target variable into classes and therefore involves a loss of information. 
More recently, new other methods have emerged by using deep learning approach as \cite{sen2023dealing}, \cite{ding2022deep} or \cite{gong2022ranksim}. \cite{yang2021} proposes to use kernel density estimates to improve learning from imbalanced data with continuous targets.

The aim of this paper is to propose a very general method, which we shall call GOLIATH (for Generalized Oversampling for Learning from Imbalanced datasets and Associated Theory) to deal with the imbalanced regression problem.  The first step of GOLIATH consists to generate synthetic data for covariates based on kernel density estimators. This data augmentation procedure can be used independently to generate synthetic data (supervised or non-supervised). A second step of GOLIATH  is concerned with the imbalanced regression where the generator model is combined with a wild-bootstrap procedure to generate target values given the synthetic covariates. The GOLIATH algorithm can be easily used in imbalanced classification where the label of the target variable remains unchanged. 
The main contributions of our paper can be summarized as follows: i) our generator model gathers in a generic form two large families of synthetic data oversampling based on perturbation and interpolation; ii) this generic form  also yields new synthetic data oversampling methods;  iii) when combined with a wild-bootstrap the generator model provides a new method for dealing with imbalanced regression iv) we empirically compared our generalized algorithm and its variants with several state-of-the-art approaches and we obtained great performance on several datasets.


The paper is organized as follows. 
In Section  \ref{proposition} we give a general form of our data augmentation procedure corresponding to the first step of GOLIATH. We study some standard perturbation and interpolation methods that are included in this approach, such as SMOTE and Gaussian Noise.  
In Section \ref{newkernel} we develop the theory to obtain new generators.
In Section \ref{specific} we will look more closely at the imbalanced regression. This is the second step of GOLIATH where we combine generators with a wild bootstrap to generate synthetic  target values. 
Section \ref{synthesis} presents an overview on the GOLIATH algorithm. Numerical results on several applications are presented in Section \ref{numeric}.

\section{A New Kernel-Based Oversampling Approach} \label{proposition}

\subsection{General Formulation}
\label{generalform}

We consider a sequence of observations $ \{(\boldsymbol{x}_1,y_1), \cdots, (\boldsymbol{x}_n,y_n)\}$, which are realizations of $n$  iid random variables  $(\boldsymbol{X},Y)$, where the target variable $Y$ is univariate and the covariate $\boldsymbol{X}$ is a $p$-dimensional random vector. The components of $\boldsymbol{X}=({X_1,\cdots,X_p})$  are supposed to be continuous or discrete and $Y$ is supposed to be either qualitative (classification) or quantitative (regression). 

Write $\tilde{\bs{x}}=\{\bs{x}_1,\cdots, \bs{x}_n\}$ the set of all observations. We propose a generalized oversampling procedure based on the form of the following weighted kernel density estimate: 
\begin{equation}
    \label{kernelForm} 
    g_{\bs{X}^*}(\bs{x}^*| \tilde{\bs{x}}) = \displaystyle \sum_{i \in {\cal I}} \omega_i K_i(\bs{x}^*,\tilde{\bs{x}}),
\end{equation}
where $(K_i)_{i \in {\cal I}}$ is a collection of kernel, $(\omega_i)_{i \in {\cal I}}$ is a sequence  of positive weights with
$\sum_{i \in {\cal I}} \omega_i=1$, and $\cal I$ represents a subset of $\{1,2,\cdots, n\}$. 
Here the index $^*$ stands for the synthetic data. 
In (\ref{kernelForm}) we propose a general form for the  conditional density for the synthetic data generators. 
The objective is to use the flexibility of the kernels to estimate the density of covariates in order to obtain synthetic data that reflects the  distribution of the observations. 


We propose to show that (\ref{kernelForm}) generalizes  perturbation-based  and interpolation-based synthetic data oversampling. 
We give an illustration with the basic algorithms ROSE, Gaussian Noise and SMOTE in Sections  \ref{revis1} and \ref{revis2} where we demonstrate that  these methods are particular cases of the generalized form (\ref{kernelForm}), with corresponding  parameters  summarized in 
Appendix.
Several existing methods can be rewritten in the form  (\ref{kernelForm})  and we give some illustrations in Appendix. 
In Section \ref{newkernel} we will show that some new methods can be deduced from the generic form (\ref{kernelForm}) and we will compare some of them  
 to current competitors in the imbalanced regression context.

\paragraph{Smoothed Boostrap Techniques}
 The generators in (\ref{kernelForm}) can be considered as  smoothed bootstrap methods   (\cite{silverman1987bootstrap}, \cite{hall1989smoothing}, \cite{de1992smoothing}). Indeed, the smoothed bootstrap consists in drawing samples from kernel density estimators of the distribution. This bootstrap can be  decomposed into two steps: first, a seed is randomly drawn and second, a random noise from the kernel density estimator is added to obtain a new sample. In the form (\ref{kernelForm}), the first step is represented by the drawing weight $\omega_i$ and the second by the kernel $K_i(x)$. 
 Convergence properties of  smoothed bootstrap are given in  \cite{de2008multivariate} and \cite{falk1989weak}. As described by the authors, the smoothed bootstrap \textit{provides better performances than classical bootstrap when a proper choice of smoothing parameters is used}. They proved the consistency of the smoothed bootstrap with classical multivariate kernel estimator and more specifically the convergence in Mallows metric. Other works have focused on the consistency of the multivariate kernel density estimate and proposed a relevant bandwidth matrix, see for instance  \cite{silverman1986density}, \cite{scott2015multivariate} and  \cite{duong2005cross}. 
 Note also that a kernel density estimator is a special case of mixture models with as many components as observations.

\subsection{Rewriting Interpolation Approaches}
\label{revis1}
 
As presented in \cite{fernandez2018smote}, the Synthetic Minority Oversampling Technique (SMOTE) \cite{Chawla2002Smote} is considered a "de facto" standard for learning from imbalanced data and has inspired
a large number of methods to handle the issue of class imbalance. It is also one of the first techniques adapted to  imbalanced target values in  regression with  \cite{torgo2013smote}. SMOTE algorithm can be summarized as follows: 
at each step of the data augmentation procedure\footnote{In the original version of SMOTE, the seed is drawn successively with a loop and not randomly. These two ways are very close when the generated sample size is large}, an observation is randomly selected, which we shall call {\it a seed}.  We will denote by    $S$ the random variable indicating the index of the seed, that is $S=i$ f the $i$th observation has been selected, and we denote by  $S(1), \cdots, S(k)$ the $k$ nearest neighbors ($k$-nn) of $\bs{x}_S$.  Given $S$, a neighbor denoted by $N_k(S)$ is randomly chosen among $S(1), \cdots, S(k)$. 
The  new data is generated by linear interpolation between $S$ and $N_k(S)$. 
 We have  $\mathbb{P}(S = i) = \frac{1}{n} ,\, \forall i=1,\cdots,n$, and 
  $\mathbb{P}(N_k(S) = S(\ell)) = \frac{1}{k} ,\, \forall \ell=1,\cdots,k$.
Finally, writing $\bs{X}^*$ the synthetic random vector we have $\bs{X}^* =  \lambda \bs{x}_S + (1-\lambda) N_k(S)$,  with $\lambda$ uniformly distributed  $\mathcal{U}([0;1])$. 

To show that this  approach is a particular case of (\ref{kernelForm}) we  proceed in three steps: 
\bit 
\item[1] Conditionally to $S$ and $N_k(S)$, the $j$th  component of $\bs{X}^*$ is generated by a uniform distribution 
{\small
\begin{align*}
g_{\bs{X}^*}^{SMOTE}\big(\bs{x}_j^*|\tilde{\bs{x}},S = i, N_k(S) = \bs{x}_i(\ell)\big) = \frac{\ind_{[\bs{x}_{ij},\bs{x}_{ij}(\ell)]}(\bs{x}_j^*)}{|\bs{x}_{ij}(\ell) - \bs{x}_{ij}|} = \frac{\ind_{[0,\bs{x}_{ij}(\ell)-\bs{x}_{ij}]}(\bs{x}_j^*-\bs{x}_{ij})}{|\bs{x}_{ij}(\ell) - \bs{x}_{ij}|},
\end{align*}
}%
where  $\bs{x}_s(\ell)$ is the $\ell${th} nearest neighbors of $\bs{x}_s$. Each component of  $\bs{X}^*$ is drawn by the same uniform variable, that is  $\bs{X}^*_j =  \lambda \bs{x}_{Sj} + (1-\lambda) N_k(S)_j$ for $j=1,\cdots, p$,  and by abuse of notation we write the multivariate generating density as follows
{\small
\begin{align*}
g_{\bs{X}^*}^{SMOTE}\big(\bs{x}^*|\tilde{\bs{x}},S = i, N_k(S) = \bs{x}_i(\ell)\big) & = \displaystyle  \frac{\ind_{[0,\bs{x}_{i}(\ell)-\bs{x}_{i}]}(\bs{x}^*-\bs{x}_{i})}{|\bs{x}_{i}(\ell) - \bs{x}_{i}|}.
\end{align*}
}%

\item[2] Conditionally to $S$, $\bs{X}^*$ is generated according  to a uniform mixture model (UMM) on the segments between $\bs{x}_S$ and its $k$-nn. The same mixture component is used for each component giving  
{\small
\begin{align*}
g_{\bs{X}^*}^{SMOTE}(\bs{x}^*|\tilde{\bs{x}},S = i)  = 
 \frac{1}{k} \sum_{\ell=1}^k  \displaystyle  \frac{\ind_{[0,\bs{x}_{i}(\ell)-\bs{x}_{i}]}(\bs{x}^*-\bs{x}_{i})}{|\bs{x}_{i}(\ell) - \bs{x}_{i}|} .
\end{align*}
}%

\item[3] More generally, $\bs{X}^*$ is generated according to a mixture of UMM as follows: \\
since $ \mathbb{P}(S = i ) = \frac{1}{ n}$, we have
{\small
\begin{align}
g_{\bs{X}^*}^{SMOTE}(\bs{x}^*|\tilde{\bs{x}})  & = \sum_{i=1}^n g^{SMOTE}(\bs{x}^*|\tilde{\bs{x}},S = i) \times \frac{1}{n}  
 = \frac{1}{n} \sum_{i=1}^n \frac{1}{k} \sum_{\ell=1}^k \displaystyle \frac{\ind_{[0,\bs{x}_{i}(\ell)-\bs{x}_{i}]}(\bs{x}^*-\bs{x}_{i})}{|\bs{x}_{i}(\ell) - \bs{x}_{i}|}   \\ 
&  =  \frac{1}{n} \sum_{i=1}^n K_i^{SMOTE}(\bs{x}^*,\bs{x})
\end{align}
}%
We finally  obtain the form (\ref{kernelForm}) with ${\cal I} =[0,n]$, $\omega_i=1/n$ and  $K_i(\bs{x}^*,\bs{x})=K_i^{SMOTE}(\bs{x}^*,\bs{x})$. In the specific case of imbalanced classification, $n$ represents the number of observations in the minority class.
\eit 
This new writing represents the  conditional SMOTE density given the observation $\tilde{\bs{x}}$. We can relate this expression to   the work of \cite{elreedy2023theoretical} in which the authors give an expression of the  unconditional SMOTE density, that is integrating the distribution over $\tilde{\bs{x}}$ in a context of class minority.  Other methods derived from SMOTE can be recovered by (\ref{kernelForm}) (see the Appendix). 



 \subsection{Rewriting Perturbation Approaches}
 \label{revis2}
We illustrate (\ref{kernelForm}) by recovering two classical data augmentation procedures as follows: 
\bit
\item
At each step of the ROSE algorithm (see \cite{Menardi2014ROSE}) the seed $S$ is selected randomly. Given $S$ a synthetic data is generated with a multivariate  density 
{\small
\begin{align*}
g_{\bs{X}^*}^{ROSE}(\bs{x}^*|\tilde{\bs{x}}, S=i) & = 
 K_{H_n}^{ROSE}(\bs{x}^*-\bs{x}_i)= \frac{1}{|H_n|^{1/2}} K(H_n^{-1/2}(\bs{x}^*-\bs{x}_i))),
\end{align*}
}%
  where   $K$ denotes the multivariate Gaussian kernel and  
$H_n =  diag(h_1,...,h_p)$ is the bandwidth matrix proposed by Bowman and Azzalini \cite{Bowman1999AppliedST}, with $h_q = (\frac{4}{(p+2)n})^{1/(p+4)} \widehat{\sigma}_q, q=1,...,p$. 
Finally, a synthetic random variable $\m X^*$ is generated with the density
{\small
\begin{align*}
g_{\bs{X}^*}^{ROSE}(\bs{x}^*|\tilde{\bs{x}}) & = 
\frac{1}{n} \sum_{i=1}^n K_{H_n}^{ROSE}(\bs{x}^*-\bs{x}_i) = \sum^n_{i=1} \omega_i K_{H_n}^{ROSE}(\bs{x}^*-\bs{x}_i)). 
\end{align*}
}%
\item
Similarly to ROSE, at each step of the Gaussian Noise algorithm  (see \cite{Lee2000GN}) a seed is selected and synthetic data is generated. Finally, the generating multivariate  density has the form 
{\small
\begin{align*}
g_{\bs{X}^*}^{GN}(\bs{x}^*|\tilde{\bs{x}}) = \frac{1}{n} \sum_{i=1}^n K_{H_n}^{GN}(\bs{x}^*-\bs{x}_i) = \frac{1}{n} \sum^n_{i=1} \frac{1}{|H_n|^{1/2}} K(H_n^{-1/2}(\bs{x}-\bs{x}_i)) ,
\end{align*}
}%
where  $H_n^{GN} = diag(h_1,...,h_p)$,  $h_q = \sigma_{noise} \w{\sigma}_q, q=1,...,p$.  
\eit    
Both cases are particular cases of (\ref{kernelForm}) 
with $\omega_i = \frac{1}{n}$ and $K_i(\tilde {\bs{x}},\bs{x}) = K_{H_n}(\bs{x}-\bs{x}_i)$, i.e. the same Gaussian kernel for all observations but with a different bandwidth matrix. 


Although there are many extensions of these methods, especially SMOTE, these techniques suffer from some drawbacks. For the interpolations techniques, the directions in the data space are limited and deterministic because they depend only on the k-nn. Moreover, the distance from the seed is also limited because the new sample is on the segment with the drawn nearest neighbor. For the perturbation techniques, the directions in the data space are randomly generated and so they can more explore the space. The distance between the new sample and the seed is also unbounded. However, the directions are randomly chosen and do not respect the correlation between the data and their support and the correlations between variables.

\section{New Kernel-Based Methods} 
\label{newkernel}
 
\subsection{Generalized Interpolation Approaches}
 \label{generalizedinterpolation}
 We propose to generalize the previous form of the SMOTE algorithm as follows: 
 {\small
 \begin{align*}
     g_{\bs{X}^*}^{int}(\bs{x}^*| \tilde{\bs{x}})  = \sum_{i \in \mathcal{I}} \omega_i K_i^{SMOTE}(\bs{x}^*,\bs{x}) = \sum_{i \in \mathcal{I}} \omega_i \sum_{\ell \in \mathcal{J}_{i}} \pi_{\ell|i} \; g^{int}_{i,\ell}(\bs{x}^*| \tilde{\bs{x}})
 \end{align*}
}%
where $g^{int}_{i,\ell}(\bs{x}^*| \tilde{\bs{x}})$ is an interpolation function  on $[\bs{x}_i, \bs{x}_i(\ell)]$, 
$\mathcal{J}_i$ denoting the set of $k$-nn associated to $\bs{x}_i$. 
It is remarkable to note that SMOTE is a particular case with  $\omega_i = \frac{1}{n}$, $\pi_{\ell|i} = \frac{1}{k}$ and $g^{int}_{i,\ell}(\bs{x}^*|\tilde{\bs{x}}) = \frac{\ind_{[0;\bs{x}_i(\ell) - \bs{x}_i]}(\bs{x}^* - \bs{x}_i)}{|\bs{x}_i(\ell) - \bs{x}_i|}$ represents a uniform distribution between the vectors $\bs{x}_i$ and $\bs{x}_i(l)$. 

\paragraph{Nearest Neighbors Smoothed Bootstrap}
Since the uniform distribution coincides with the  Beta distribution with parameters $\alpha=\beta=1$, a natural extension of SMOTE is to consider a more general Beta distribution. We find the same idea in \cite{yao2022c} within another context. The very flexibility of the Beta  distribution  suggests us to 
propose 
$\bs{X}^* =  \lambda \bs{x}_{i} + (1-\lambda) \bs{x}(\ell)_i$ for $i=1,\cdots, n$, where $\lambda$ follows a generalized Beta distribution. By abuse of notation, we get the following   interpolation function: 
 {\small
 \begin{align*}
g^{int}_{i,\ell}(\bs{x}^*|\tilde{\bs{x}}) = \displaystyle \frac{\Gamma(\alpha + \beta)}{\Gamma(\alpha)\Gamma(\beta)} (\bs{x}^*-\bs{x}_{i})^{\alpha-1}(\bs{x}_{i}(\ell)-\bs{x}^*)^{\beta-1} \ind_{[0;\bs{x}_{i}(\ell) - \bs{x}_{i}]}.
 \end{align*}
}%

We emphasize  that the extensions of SMOTE in the literature can replace the original SMOTE by this generalization. This provides more flexibility and a potential improvement of the results. 

\paragraph{Extended Nearest Neighbors Smoothed Bootstrap} 
Finally, the previous methods based on interpolation are limited to the "seed - $k$-nn segments" and therefore do not reach all the data space. To avoid generating on a bounded or discrete support we propose to  extend these approaches to any part of the support by adding a Gaussian distribution on the segment as follows:  
 {\small
 \begin{align*}
     g^{e-int}(\bs{x}^*| \tilde{\bs{x}}) = \sum_{i \in \mathcal{I}} \omega_i \sum_{\ell \in \mathcal{J}_i} \pi_{\ell|i} \; g^{e-int}_{i,\ell}(\bs{x}^*| \tilde{\bs{x}})
 \end{align*}
}%
where the extended interpolation function is a Beta Gaussian mixture, that is,   $g^{e-int}_{i,\ell}(\bs{x}^*| \tilde{\bs{x}})$ is the density of a Gaussian distribution $N(\theta,\sigma^2)$ where $\theta$ is generated by 
$g^{int}_{i,\ell}(\bs{x}|\tilde{\bs{x}})$. To rely on the recent literature, we remark that SASYNO algorithm \cite{gu2020self} is a special case of this methodology. This extended version can be viewed as  a hybrid method between interpolation and perturbation techniques. It provides a good compromise between the interpolation and perturbation approaches because it can generate in the whole data space as the perturbation approach i.e. constraint-free, but assigns a distribution to the directions towards the segments, that is orienting the perturbation toward the $k$-nn.   

We also tried to adapt the k-nearest neighbors density estimate \cite{biau2015lectures} that is a bandwidth-variable kernel (also called a balloon kernel) as a generator but its computation time is currently too high to be used.

\subsection{Generalized Perturbation Approaches}
  \label{generalizedperturbation}

\paragraph{Classical Smoothed Bootstrap}
As the ROSE and GN techniques use a multivariate Gaussian kernel estimate with a diagonal bandwidth matrix, we can rewrite their associated generating density as follow 
{\small
 \begin{align}
 \label{ROSEGN}
     g_{\bs{X}^*}(\bs{x}^*|\tilde{\bs{x}} ) & 
     = \sum_{i=1}^n \omega_i \prod_{j=1}^p  K_{h_j}(x^{*}_j-{x}_{ij})
 \end{align}
}%
with $K_{h_j}(u) = (2 \pi)^{-1/2} h_j^{-1} e^{-\frac{1}{2 {h_j}^2} u^2}$ the univariate gaussian kernel density estimator with smoothing parameter $h_j$. 
Such kernels are clearly not adapted for asymmetric, bounded  or discrete variables. This remark is also true for the work of \cite{yang2021} which uses some symmetric kernels to improve the learning of imbalanced datasets. Another remark about  this work is the division of the target variable support into $B$ groups that involve a loss of information. 

\paragraph{Non-Classical Smoothed Bootstrap}
To fix the drawback of the classical kernel we extend (\ref{ROSEGN})  by   
adapting (\ref{kernelForm}) to the  support of $\bs{x}$, considering some non-classical kernels  (we refer to some works handling the kernel density estimation for specific distributions inspired from \cite{bouezmarni2010nonparametric}, \cite{some2015estimations}, \cite{hayfield2008nonparametric}, \cite{chen2000probability}). We suggest rewriting the form   (\ref{kernelForm}) as  
{\small
 \begin{align*}
     g_{\bs{X}^*}^{per}(\bs{x}^* | \tilde{\bs{x}}) & = \sum_{i \in \mathcal{I}} \omega_i \prod_{j=1}^p  K_{h_j}(x^{*}_j,{x}_{ij})
 \end{align*}
}%
where $K_{h_j}(u,x)$ is a univariate kernel adapted to the nature of the $j$th variable and specifically defined on $x$ as follows: 
\begin{itemize}
    \item Gaussian kernel for a variable defined on $\mathbb{R}$ (classical kernel):  
    {\small
     \begin{align*}
     K_{h}(u,x) = \frac{1}{h\sqrt{2 \pi}} e^{-\frac{1}{2} (\frac{u-x}{h})^2}.
     \end{align*}
     }%
    \item Binomial kernel for a discrete variable defined on $\mathbb{N}$:
    {\small
     \begin{align*}
     K_{h}(u,x) = \frac{(x+1)!}{u!(x+1-u)!} \left( \frac{x+h}{x+1} \right)^{{u}} \left( \frac{1-h}{x+1} \right)^{x+1-u}.
     \end{align*}
     }%
    
    \item Gamma kernel for a positive asymmetric distribution defined on $[a,+\infty]$:
    {\small
     \begin{align*}
     K_{h}(u,x) = \frac{{u}^{(x-a)/h}}{\Gamma(1+(x-a)/h)h^{1+(x-a)/h}}\exp \big(\frac{-u}{h}\big) \ind_{[a,+\infty]}(u).
     \end{align*}
     }%
    
    \item Negative Gamma kernel for a negative asymmetric distribution defined on $[-\infty,b]$:  
    {\small
     \begin{align*}
     K_{h}(u,x) = \frac{{u}^{-(x-b)/h}}{\Gamma(1-(x-b)/h)h^{1-(x-b)/h}}\exp \big(\frac{-u}{h}\big) \ind_{[-\infty,b]}(u).
     \end{align*}
     }%
    
    \item Beta kernel for a variable defined on $[0,1]$:
    {\small
     \begin{align*}
     K_{h}(u,x) = \frac{{u}^{x/h}(1-u)^{(1-x)/h}}{\mathcal{B}(\frac{x}{h}+1,\frac{1-x}{h}+1)} \ind_{[0,1]}({u}).
     \end{align*}
     }%
    
    \item Truncated Gaussian kernel for a variable defined on $[a,b]$: 
    {\small
     \begin{align*}
     K_{h}(u,x) = \frac{\alpha}{h\sqrt{2 \pi}} e^{-\frac{1}{2} (\frac{u-x}{h})^2} \ind_{[a,b]}(u), \alpha := \left( \int_a^b \frac{1}{h\sqrt{2 \pi}} e^{-\frac{1}{2} (\frac{u-x}{h})^2} \right) ^{-1}
     \end{align*}
     }%
    
\end{itemize}

Note that if the Dirac kernel ($\ind_{\bs{x}=\bs{x}_i}$) is used, we get the standard bootstrap: \ref{kernelForm} includes also the simple oversampling. It is important to note that the GOLIATH algorithm uses an estimation of the smoothing parameter $h$ provided by some specific R-package dedicated to the density estimation (for instance, it uses the Silverman estimation for the Gaussian kernel). Their estimates are based on properties of univariate consistency. Another technique to deal with skewed or heavy-tailed distributions is to apply a transformation of the data in order to use classical kernel density estimation \cite{charpentier2015log}, \cite{charpentier2010beta} but it necessitates proposing a relevant transformation. 

\textbf{Remark:} The  use of a diagonal bandwidth matrix in (\ref{ROSEGN}) does not take into account the correlation between variables. To improve this issue, we could consider a full (symmetric positive definite) smoothing matrix. In that case, we would use a multivariate kernel density estimate considering the correlation between the variables which would be optimal for generating data. However, the estimation of this kind of matrix is usually very difficult and expensive, the thesis of Duong \cite{duong2004bandwidth} and his associated R-package \cite{duong2007ks} proposed a Multivariate Gaussian Kernel and a maximum of 6 dimensions. These works, despite their high quality, are very limited in practice since on the one hand datasets contain generally more than 6 variables, and on the other hand Gaussian Kernel estimators are  inappropriate for mixed data.

\subsection{Goliath Overview}
\label{synthesis}


  \begin{minipage}{\linewidth}
      \centering
      \begin{minipage}[c]{0.45\linewidth}
            \begin{algorithm}[H]
            \begin{algorithmic}
            \scriptsize
                \STATE {\bfseries Input} covariate X; target variable Y=null; mod="mix"; type; method-Y=1; clustering=F; seed s=null; components GMM m=n-row(X); weights w=rep(1,n-row(X)); synthetic data sample size N=n-row(X); parameter p)\\
                {\bf Clustering}  \hskip1em// \textit{Optional application of a clustering on the train}
                \STATE clust = Cluster(X,Y,clustering) \hskip1em// \textit{clust=1 for all samples if clustering=F}\\
                {\bf Seed drawing} \hskip1em// \textit{weighted oversampling}
                \STATE if s = null then s = draw(X,N,w)\\
                {\bf X generation} \hskip1em// \textit{Synthetic data generation for the covariates}
                \STATE if m < n then synth = GMM(X,Y,m,N) 
                \STATE \hskip1em else for each c in clust: 
                \STATE \hskip2em if type = "CSB" then synth = G-CSB(X,N,w, s, p)  
                \STATE \hskip2em else if type = "NCSB" then synth = G-NCSB(X,N,w, s, p) 
                \STATE \hskip2em else if type = "NNSB" then synth = G-NNSB(X,N,w, s, p) 
                \STATE \hskip2em else if type = "eNNSB" then synth = G-eNNSB(X,N,w, s, p)    \\ 
                \hskip2em{\bf Y generation} \hskip1em// \textit{Optional Synthetic Y generation}
                \STATE \hskip2em if Y <> null then synth[,Y] = G-Y(X,Y,synth,s,method-Y, p)\\
                \STATE \hskip1em end For \\
                {\bf Output} 
                \STATE if m<n or mod = "synth" then return synth \\
                \STATE else if mod = "augment" then return (X,Y) + synth \\
                \STATE else return mix((X,Y),synth,s) 
            \label{algoGOLIATH}
            \end{algorithmic}
            \end{algorithm}
      \end{minipage}
      \hspace{0.05\linewidth}
      \begin{minipage}[c]{0.45\linewidth}
        \begin{figure}[H]
                \includegraphics[width=\textwidth]{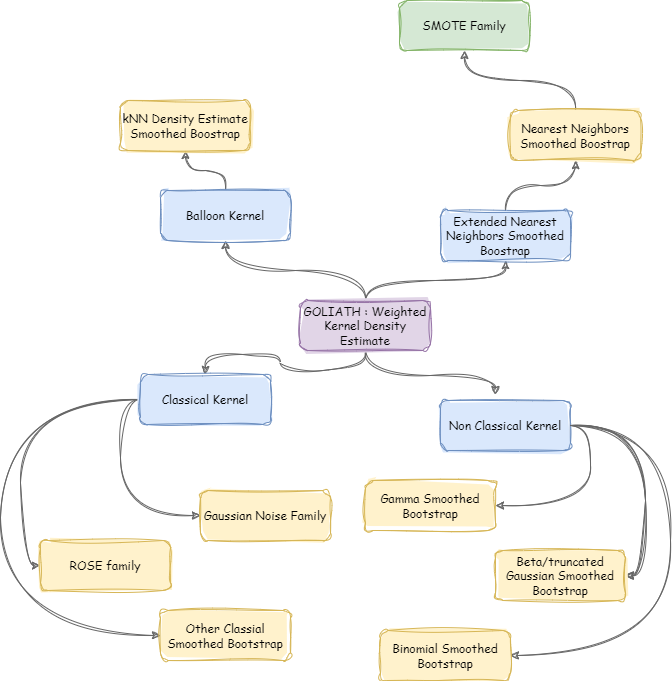}
                \caption{GOLIATH cartography}
                \label{carto}
        \end{figure}
      \end{minipage}
  \end{minipage}

The GOLIATH algorithm is summarized in Figure \ref{algoGOLIATH}. A cartography of GOLIATH is also given in Figure \ref{carto}.
The algorithm gives the possibility to choose, with the "mod" parameter, the kind of the returned sample: a full synthetic dataset, an augmented one, or a mixed one. The mixed sample is constructed as follows: keep the original observation for the first occurrence of the seed and synthetic data for the next. This mode corresponds to performing an undersampling and an oversampling. More precisely, to preserve the maximum of information and avoid potential overfitting, we suggest to: keep the initial observation for its first drawing and generate synthetic data from it for the other drawing which is the "mix" mod in the GOLIATH algorithm. More details on this option are given in Appendix.

 \section{GOLIATH as a Solution for Imbalanced Regression}
 \label{specific}

Using methods proposed in 2.3.1 and 2.3.2 we can generate  synthetic covariates $X^*$. 
We then  have to generate   the target variable $Y$   given $X^*$ obtained from the first generator step of GOLIATH.  
In this sense we propose here to define the drawing weights $\omega_i$ as the inverse of the kernel density estimate for the target variable $Y$: the more isolated an observation is, the higher its drawing weight. The same idea is proposed in \cite{steininger2021density}. We also defined some safeguards to avoid getting some weights too high. 

Finally, the target variable is not generated in the same way as the covariate. Once the covariates are generated from our generator models, we propose to adapt a wild-bootstrap technique\footnote{The kernel regression (Nadaraya-Watson estimator) was also tested but not selected because its high computation time and poor performance} \cite{wildBoostrap} as follows: 
i) train a Random Forest on the initial sample; ii) predict  target variable  $\w y_i$; iii) draw uniformly a prediction error $\epsilon_k$ to generate a new $y_i^* := \w y_i + \epsilon_k v_i$  where $v_i$ is a random variable. We suggest an adaptation of this method to synthetic data in considering the impact of getting new covariate  $y_i^* := \w y_i + |\epsilon_k|v_i \times sign(\w{y_i}-\w{y_i^*})$. This form is close to the Wild Bootstrap version with the Rademacher distribution. This is the second step of GOLIATH. Other methods to generate $Y$ are proposed in Appendix 
(giving lower performances in our applications). 
The idea behind this proposition is to consider the prediction error and the impact of the synthetic covariate on the target variable. The choice of using a random forest is justified by its good predictive performance, its non-parametric nature, and the possibility of getting an error distribution for a same value of the target variable.

\section{Application in Imbalanced Regression}
\label{numeric}

Although the GOLIATH algorithm can be applied for the classification tasks, we focus on the imbalanced regression context because of the natural capacity of the form (\ref{kernelForm}) to handle continuous variables. 

For each dataset, we construct a test sample as 10-30\% of the initial dataset and use the weighting previously defined: the inverse of the kernel density estimate for the target variable $Y$. In this way, we obtain a test sample with a distribution of the target variable $Y$ approximately uniform. We construct an artificial imbalanced dataset from the remaining sample. More details on the protocol are given in Appendix.
One of the objectives of our approach is to obtain better performance than the imbalanced train dataset. We compare our results to existing methods to deal with imbalanced regression from the \textit{UBL} R-package, \cite{branco2016ubl},: classical oversampling, SMOTE, Gaussian Noise, SMOGN, WERCS and ADASYN from the python-package \textit{ImbalancedLearningRegression} \cite{wu2023imbalancedlearningregression}. These techniques are used with their automatic relevance function and the same parameters as GOLIATH if any, in particular $k$ for SMOTE and $pert$ for the Gaussian Noise.

To avoid sampling effects and obtain a distribution of  prediction errors we ran 10 train-test datasets.  In the same way, to avoid getting results dependent on some learning algorithms we use 10 models from the \textit{autoML of the H2O R-package} \cite{H2OAutoML20} among the following algorithms: Distributed Random Forest, Extremely Randomized Trees, Generalized Linear Model with regularization, Gradient Boosting Model, Extreme Gradient Boosting and a Fully-connected multi-layer artificial neural network. We present here the aggregated results of the models, a more detailed analysis is available in Appendix.

We then compute the following metric: RMSE and MAE, weighted-RMSE with our weighting function giving more importance to the rare values, and correlation between predictions and $Y$ values of the test sample. Since the test sample is balanced on the target variable, we considered here the RMSE and MAE metrics as relevant to provide an overview of the average error across the whole target variable. We can find the same metrics in \cite{ren2022balanced} and \cite{yang2021}.
\paragraph{Remarks on GOLIATH.} The basic methods (Gaussian Noise or SMOTE) are different from the UBL version because i) the generation of the target variable $Y$ is realized with wild bootstrap and considers the new synthetic attributes and ii) the weights $\omega$ is defined for all samples while UBL uses a relevant function that divides the dataset into rare and frequent sets. Like the ROSE and Gaussian Noise algorithms, GOLIATH takes into account a parameter tuning the level noise for perturbations approaches (description in Appendix). It is also possible to use a clustering (Gaussian Mixture Model) in GOLIATH in order to apply a generation by cluster.  All datasets provided by GOLIATH in the applications were provided with the mod "mix". Note that the ROSE algorithm did not exist for the imbalanced regression.


\subsection{Illustrative Application} \label{illu_appli}

In order to get a reference for predictive performance, we have chosen to use a balanced dataset from which we build an imbalanced train dataset. The dataset, named \textit{
SML2010 Data Set} is available on the Machine Learning repository UCI\footnote{\url{https://archive.ics.uci.edu/ml/datasets/SML2010}}. It is composed of 24 numeric attributes and 4137 instances. The target variable is the indoor temperature (we construct a unique target variable as the mean temperature of dinning-room and the temperature of the room). We train, with the autoML, the following train dataset: 
\bit
\item Reference values: Full sample (\textit{FTrain}), Imbalanced (\textit{Imb})
\item Benchmark: UBL-Oversampling (\textit{UBL-OS}), UBL-SMOTE for regression (\textit{UBL-SMOTE}), UBL-Gaussian Noise for regression (\textit{UBL-GN}), UBL-SMOGN for regression (\textit{UBL-SMOGN}), UBL-WERCS (\textit{UBL-WERCS}), IRL-ADASYN (\textit{IRL-OS})
\item GOLIATH (step 1): Oversampling (\textit{G-OS}), Gaussian Noise (\textit{G-GN}), Gaussian Noise with GMM-clustering (\textit{G-GNwCl}), ROSE (\textit{G-ROSE}), ROSE with a GMM-clustering (\textit{G-ROSEwCl}), SMOTE (\textit{G-SMOTE}), Non classical Smoothed Boostrap with contraints on the distributions (\textit{G-NCSB}), Classical Smoothed Boostrap (\textit{G-CSB}), Nearest Neighbors Smoothed Bootstrap, with Beta distribution, (\textit{G-NNSB}), Nearest Neighbors Smoothed Bootstrap  with k-NN weights proportionates to the distance from the seed (\textit{G-NNSBw}), Extended Nearest Neighbors Smoothed Bootstrap (\textit{G-eNNSB}). 
\eit

\begin{figure}[ht]
\centering
\begin{subfigure}{0.49\textwidth}
    \includegraphics[width=1\textwidth]{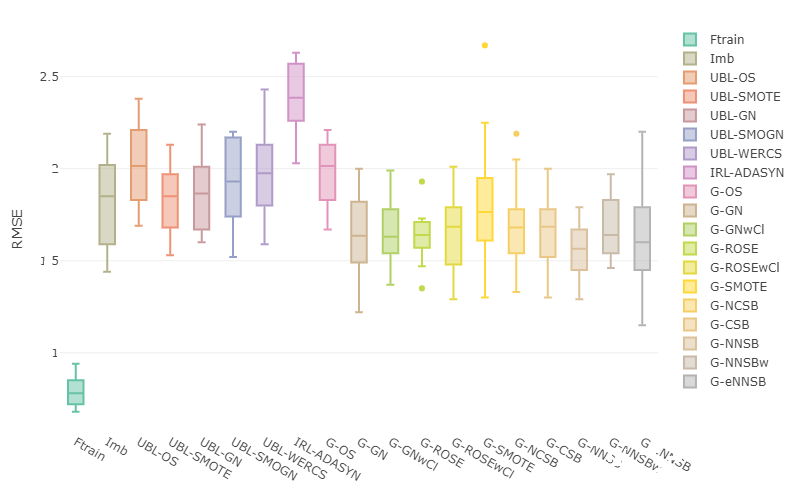}
    \caption{RMSE Boxplots}
    \label{RMSE_AppliIllu}
\end{subfigure}
\hfill
\begin{subfigure}{0.49\textwidth}
    \includegraphics[width=1\textwidth]{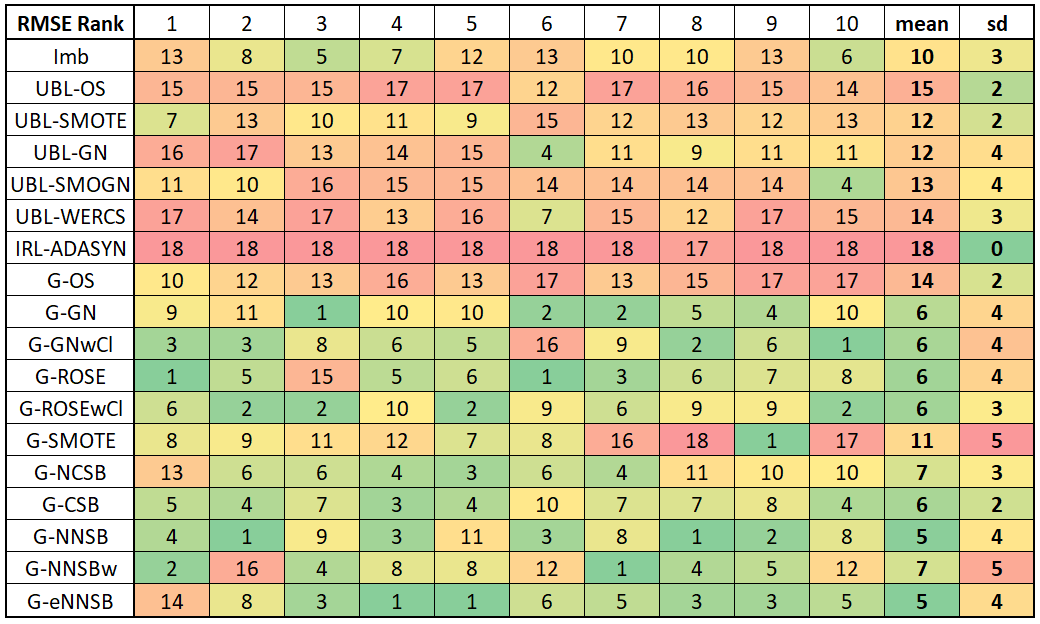}
    \caption{RMSE Ranking}
    \label{RMSERank_AppliIllu}
\end{subfigure}
\caption{Boxplots of RMSE and RMSE-rank for all train samples}
\end{figure}

Figure \ref{RMSE_AppliIllu} shows the results for RMSE on the test sample, with $Y$ quite uniformly distributed. The weighted-RMSE and MAE metrics are shown in Appendix 
and present similar results. We can observe that the GOLIATH algorithm presents an RMSE smaller than the imbalanced sample and than the benchmark techniques, whatever the generators. The GOLIATH-oversampling is comparable to the UBL-oversampling which confirms, on this dataset, the relevance of the weighting.  
The Non-Classical Smoothed Bootstrap (NCSB) is as efficient as the Classical Smoothed Bootstrap (CSB, ROSE, GN). However, it provides realistic values for the variables. We can see some examples of inconsistency in Appendix. The results show that the clustering seems improve the performance.
The Nearest Neighbors Smoothed Bootstrap (NNSB) and Extended Nearest Neighbors Smoothed Bootstrap(eNNSB) outperform the original SMOTE. It is important to note that the different parameters (the number of nearest neighbors for interpolation techniques and the level of noise for perturbation techniques) are arbitrary and not optimized here. 
The heatmap in Figure \ref{RMSERank_AppliIllu} shows the robustness of the methods with their rank by run with respect to the RMSE. We can observe, based on the mean and standard deviation of the rank,  that the Classic Smoothed Bootstrap, the Non-Classical Smoothed Bootstrap and the Nearest Neighbors Smoothed Bootstrap are the best approaches here.  

The RMSE-rank represents the ranking of approaches according to the RMSE for a run: rank 1 corresponds to the training dataset that offers the smallest RMSE on the test sample. 

\subsection{Imbalanced Regression Applications} \label{applis}

We test our approach on several real data set from a repository provided as a benchmark for imbalanced regression problems\footnote{\url{https://paobranco.github.io/DataSets-IR/}} and presented in \cite{branco2019pre} (descriptions in Appendix). Figure \ref{figure2} presents RMSE gain (wrt the imbalanced dataset) and the median of the RMSE ranking. We can observe on these datasets that the GOLIATH algorithm empirically outperforms the state-of-the-art techniques, especially the Non-Classical Smoothed Bootstrap and the Extended Nearest Neighbors Smoothed Bootstrap.

\begin{figure}[H]
\centering
\begin{subfigure}{0.49\textwidth}
    \includegraphics[width=1\textwidth]{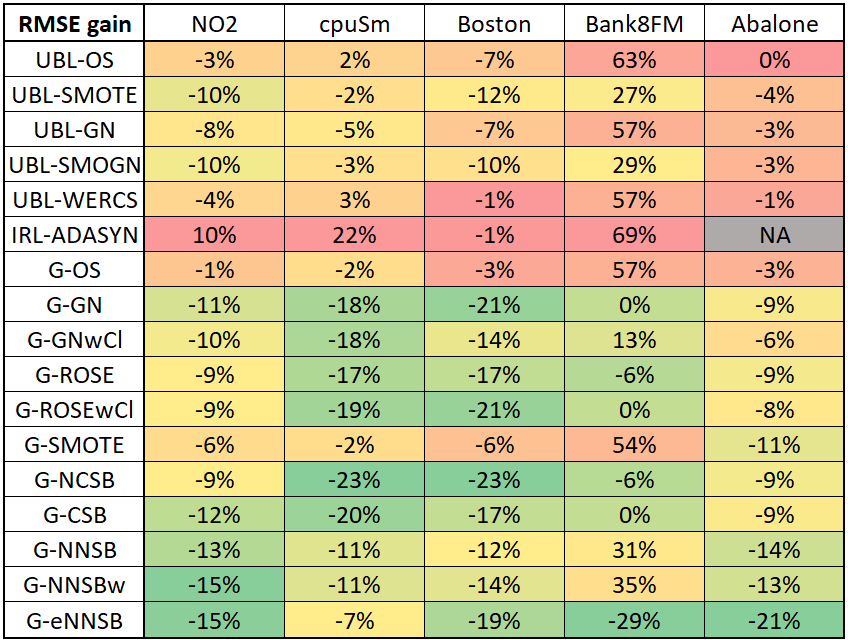}
    \caption{RMSE gain}
    \label{RMSE_gainApplis}
\end{subfigure}
\hfill
\begin{subfigure}{0.49\textwidth}
    \includegraphics[width=1\textwidth]{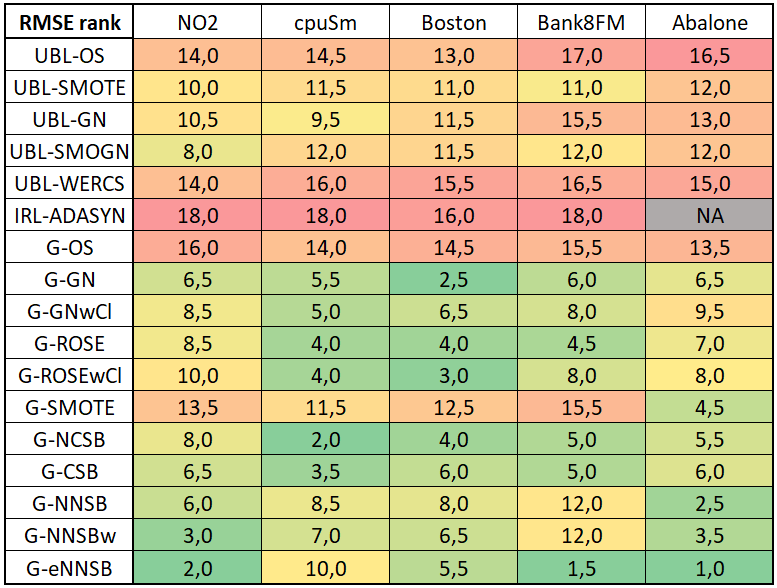}
    \caption{Median of the RMSE-rank}
    \label{RMSE_rankApplis}
\end{subfigure}
\caption{RMSE-gain and median of the RMSE-rank on the Imbalanced Regression Datasets}
\label{figure2}
\end{figure}

We can see on these several applications, with several runs, several learning algorithms, and several performance metrics that the GOLIATH approach seems relevant to deal with imbalanced regression.

\section{Discussion and Perspectives}

GOLIATH is an algorithm gathering two large families of synthetic data oversampling. Many methods can be rewritten as particular cases of it. This approach gets the advantage to obtain a general form for the generator which is based both on the theoretical foundations of kernel estimators  and  classical smoothed bootstrap techniques. It provides a general expression  for the conditional density of the  generator. 
The use of  well-chosen  kernels  makes it possible to take into account  the nature of the covariates:  continuous, discrete, totally or partially bounded. Our approach generalizes the SMOTE algorithm by providing  weights and flexible densities for interpolation. We also extend this technique  to wider support than that of the observations  by combining  interpolation and perturbation approaches.  Numerical applications in imbalanced regression models  demonstrate that GOLIATH and its variants are very competitive, especially when the generator used in step 1  is the extended nearest neighbors smoothed bootstrap.

The weights $\omega_i$ (and $\pi_{j|i}$ in the interpolation case) offer a large flexibility to use the method. For instance, it is possible to handle classification tasks by conditioning with the minority class. We could deal with multi-class classification too. It is also possible to combine  some extensions of SMOTE that propose  to focus on specific samples in the synthetic data generation (as ADASYN) with a kernel approach in order to perform the methodology.


As a perspective, a natural extension of this work is to   automate the choice of the kernel estimators, the weights, as well as some parameters according to the data. For example  by defining a weights function for the nearest neighbor instead of defining the parameter $k$. Indeed, 
the parameter $k$ is sometimes unsuitable and we could suggest a dynamic weighting depending on the neighborhood. It is also possible to define a kernel according to the neighborhood into the same dataset. 
We also could define $\omega_i$ in order to generate to a target distribution as done in \cite{stocksieker2023data}. Finally, by nature, the perturbation approaches with a diagonal bandwidth matrix do not consider the correlation between covariates. The interpolation approaches consider it but the generation is limited to the segments. The extended-SMOTE proposes a first solution. Another direction for further investigations would be to better consider the correlations between variables and be able to handle mixed data.

\newpage
\section{Supplementary Material}

\subsection{Summary of Existing Methods}

\subsubsection{Summary of the Original Methods}
\paragraph{Randomly Over Sampling Examples (ROSE):}\label{summaryROSE}

The idea of ROSE algorithm (\cite{Menardi2014ROSE}) for classification can be summarized as follows:  
\begin{enumerate}
    \item Select $y^* = \mathcal{Y}_j, j=0,1 $ with probability $1 / \pi_j$, $\pi_j = 1/2$ for binary classification. 
    \item Draw uniformly  a seed $(\bs{x_i},y_i)$ such that $y_i = y^*$, with probability $p_i= 1 / n_j$, $n_j < n$ being the size of $\mathcal{Y}_j$  
    \item Generate a sample $\bs{x}^*$ from $K_{H_j}(\cdot,\bs{x_i})$, where  $K_{H_j}$ is a probability distribution centered at $\bs{x_i}$ and with covariance matrix $H_j$.
\end{enumerate}

The authors consider  Gaussian Kernels for $K$ with diagonal smoothing matrices $H_j = diag(h_1^{j},\cdots,h_p^{j})$. Thus, the generations of new samples from the class $\mathcal{Y}_j$ correspond to data from the kernel density estimate  $f(\bf{x}|\mathcal{Y}_j)$. Under the assumption that the true conditional density underlying the data follows a Gaussian distribution, they suggest using the proposition of \cite{Bowman1999AppliedST}: \\ $h_q^{(j)} = (4/((d+2)n)^{1/(d+4)} \w{\sigma_q^{(j)}} \, (q=1,\cdots,p; j=0,1)$, where $\w{\sigma_q^{(j)}}$ is the sample estimate of the standard deviation of the q-th dimension of the observations belonging to the class $\mathcal{Y}_j$.

\paragraph{Gaussian Noise (GN):}\label{summaryGN}

The idea of GN algorithm for classification (\cite{Lee2000GN}), can be summarized as follows:
\begin{enumerate}
    \item Choose $m$ being the number of replicates of the training data from the imbalanced class that is when $y = y_{min}$.
    \item For each $\bs{x}_i$ from the imbalanced class, generate $m$ noisy replicates $\bs{x}^*$ of  the form $(\bs{x}+\bs{\epsilon},y_{min}), \bs{\epsilon} = (\epsilon_1,\cdots,\epsilon_p)$ where $\epsilon_j, j=1,\cdots,p$ is a Gaussian noise, i.e. $ x^*_q = x_q + \mathcal{N}(0,\w{\sigma}_q \times \sigma_{noise})$,  where $x_q$ is the feature value $q$ from the original observation, and $\sigma_{noise}$ is defined by the user.
\end{enumerate}

\paragraph{Synthetic Minority Oversampling Technique (SMOTE):}\label{summarySMOTE}

The idea of the SMOTE algorithm (\cite{Chawla2002Smote}) can be summarized as follows: 
\begin{enumerate}
    \item Compute the $k-$ nearest neighbors for each minority sample
    \item Do a loop on the minority class samples: generate $N$ synthetic samples for each seed $x_s$ as follows:
    \begin{enumerate}    
        \item Choose one of the $k-$ nearest neighbors of $x_s$, say $x_j(s)$. 
        \item Compute $\lambda$ as a random number between 0 and 1.
        \item Create a new synthetic sample $x^*$ defined as: $x^* := \lambda x_s + (1- \lambda)(x_j(s)-x_s)$.
    \end{enumerate}
\end{enumerate}


\subsubsection{Original Methods within the GOLIATH Form}

Table \ref{table1} summarizes the original perturbation approaches ROSE and GN and the original interpolation approaches SMOTE as the form \ref{kernelForm}. $n$ represents here the number of samples in the minority class.

{\small
\begin{table}[H]
\caption{Summary of the parametrization of the original methods within the GOLIATH form} 
\begin{center}
\begin{tabular}{ c | c | c| c | m{5cm}  } 
  Generator & ${\cal I}$ & $\omega_i$ & $K_i(x)$ & Precision  \\
  \hline
  ROSE & $[1,n]$ & $1/n$ & $\frac{1}{|H_n|^{1/2}} K(H_n^{-1/2}(\bs{x}-\bs{x}_i)))$ & $K(\cdot)$ the multivariate Gaussian kernel, $H_n$ the bandwidth matrix proposed by \cite{Bowman1999AppliedST} \\ 
  \hline
  GN & $[1,n]$ & $1/n$ &  $\frac{1}{|H_n|^{1/2}} K(H_n^{-1/2}(\bs{x}-\bs{x}_i)))$ 
& $K(\cdot)$ the multivariate Gaussian kernel, $H_n$ a diagonal matrix with a fraction of the empirical standard deviations\\ 
  \hline
  SMOTE & $[1,n]$ & $1/n$ & $ \frac{1}{k} \sum_{\ell=1}^k \displaystyle \frac{\ind_{[0,\bs{x}_{i}(\ell)-\bs{x}_{i}]}(\bs{x}^*-\bs{x}_{i})}{|\bs{x}_{i}(\ell) - \bs{x}_{i}|} $ & $K(\cdot)$ a Uniform Mixture Model with $k$ components having the same weight, depending of the $k$-nn of $\bs{x}_i$ \\  
\end{tabular}
\label{table1}
\end{center}
\end{table}
}%

\subsubsection{Other Existing Methods within the GOLIATH Form}
\label{paramExistingMethods}

Here we present some extensions of SMOTE that can be written within a simplified GOLIATH form \ref{kernelForm}. These methods are applied in Imbalanced classification. $n$ represents the number of samples in the minority class. We note by $K_i^{SMOTE}(x)$ the SMOTE kernel defined on Table \ref{table1} $ \frac{1}{k} \sum_{\ell=1}^k \displaystyle \frac{\ind_{[0,\bs{x}_{i}(\ell)-\bs{x}_{i}]}(\bs{x}^*-\bs{x}_{i})}{|\bs{x}_{i}(\ell) - \bs{x}_{i}|} $. We note by $n_{\cal I}$ the number f instances which respects the condition of $\cal{I}$. Note that it is possible to write within the GOLIATH form \ref{kernelForm} the under-sampling methods as well as over-sampling methods and hybridization.

\begin{table}[ht]
\caption{Summary of the parametrization of some extensions SMOTE methods within the GOLIATH formulation} 
{\scriptsize
\begin{center}
\begin{tabular}{  m{2cm} | m{2cm} | m{2cm} | m{2cm} | m{3cm}  } 
  Generator & ${\cal I}$ & $\omega_i$ & $K_i(x)$ & Precision  \\
  \hline
  Borderline-SMOTE (\cite{han2005borderline}) & $x_i$ such as $\frac{k^{mj}(x_i)}{k} \geq 50\%$  &  $1/n_{\cal I}$ & $K_i^{SMOTE}(x)$ & $k^{mj}(x_i)$ represents the number of majority examples among the k nearest neighbors of $x_i$, $n_{\cal I} := \sum_i \ind_{k^{mj}(x_i)/k \geq 50\%}$ \\ 
  \hline
    Safelevel-SMOTE (\cite{bunkhumpornpat2009safe}) & $x_i$ according to $SLR_{ij}$ and $SF_i$: cf below 
    & $1/n_{\cal I}$ & adapted kernel: cf below & $SFR_{ij}:=\frac{SL_{i}}{SL_{ij}}=\frac{SL(x_i)}{SL(x_j(i))}$ with $SL(x)$ the number of minority instances in k nearest neighbours for x \\ 
    Safelevel-SMOTE & $x_i$ such as $SFR_{ij}=\infty \, \& \, SL_i \ne 0$  
    &  $1/n_{\cal I}$ & $x$ (oversampling) & oversampling if $x_j(i)$ is considered as noise \\ 
    Safelevel-SMOTE  & $x_i$ such as $SFR_{ij}=1 \, \& \, SL_i=SL_{ij} $  &  $1/n_{\cal I}$ & $K_i^{SMOTE}(x)$ & SMOTE if $x_j(i)$  is considered as safe\\ 
    Safelevel-SMOTE  & $x_i$ such as $SFR_{ij}>1 $  &  $1/n_{\cal I}$ & $K_i^{SMOTE}(x)$ with $\lambda \sim \mathcal{U}([0,0.5])$ &  SMOTE closer to $x_i$ if $x_j(i)$  is considered as not safe \\     
    Safelevel-SMOTE  & $x_i$ such as $SFR_{ij}<1 $  &  $1/n_{\cal I}$ & $K_i^{SMOTE}(x)$ with $\lambda \sim \mathcal{U}([0.5,1])$ &  SMOTE closer to $x_j(i)$ if $x_i$ is considered as not safe\\    
  \hline
  ADASYN (\cite{he2008adasyn}) & $[1,n]$  &  $ \widehat{r_i} := \frac{r_i}{\sum_i r_i}$ and $r_i = \frac{\Delta_i}{k}$ & $K_i^{SMOTE}(x)$ & $\Delta_i$ represents the number of examples in the $k$ nearest neighbors of $x_i$ that belong to the majority class \\ 
  \hline
  Kernel-ADASYN (\cite{tang2015kerneladasyn}) & $[1,n]$  &  $ \widehat{r_i}:= kde(r_i)$ and $r_i = \frac{\Delta_i}{k}$ & $K_i^{SMOTE}(x)$ & evolution of the ADASYN algorithm by using a Gaussian kernel density estimate to normalize $r_i$ \\ 
  \hline
  SMOTE-TomekLink (\cite{batista2004study}) & $[1,n]$  &  1 if $x_i$ and $x^*$ do not form a Tomek link &  $K_i^{SMOTE}(x)$ & This a acceptance-rejection method according to the \textit{Tomek-link}(\cite{tomek1976two}) applied after the generation of $x^*$ with SMOTE \\ 
  \hline
  SMOTE-ENN (\cite{batista2004study}) & $[1,n]$  & 1 if $y_i = y_j(i), j \in [1,k_{ENN}]$ &  $K_i^{SMOTE}(x)$ & Similar to SMOTE-TomekLink but by using the rule of ENN (\cite{wilson1972asymptotic}), $k_{ENN}$ represents the number of the NN considering for the ENN rule \\ 
  \hline
  Kmeans-SMOTE (\cite{douzas2018improving}) & $x_i \in$ filtered clusters  & sampling weight based on its minority density into the filtered cluster  &  $K_i^{SMOTE}(x)$ & SMOTE applied on the filtered clusters defined as imbalanced cluster \\ 
  \hline
  
\end{tabular} 
\label{table10}
\end{center}
}%
\end{table}

\subsection{Differences with \textit{Utility-Based Learning} Approach}
Here we present some differences with the \textit{Utility-Based Learning} approach proposed in \cite{branco2016ubl} and  associated works, for example \cite{torgo2013smote}, \cite{branco2017smogn},  \cite{branco2019pre}, \cite{ribeiro2020imbalanced}. This is the first and main solution in the Imbalanced Regression. These works are considered references for Imbalanced Regression Learning.

\begin{table}[H]
\caption{Differences between the UBL approach and the GOLIATH approach} 
\begin{center}
\begin{tabular}{ | m{2cm} || m{5cm}| m{5.5cm} | } 
  \textbf{Characteristic} & \textbf{UBL approach} & \textbf{GOLIATH approach} \\ 
  \hline
  Rebalancing & Using a binarization of $y$ & Using the continuous distribution of $y$ \\
  \hline
  Flexibility & Limited to some adaptations of imbalanced classification methods & Adaptation of imbalanced classification methods (SMOTE family) and kernel-based methods, for continuous distributions \\
  \hline
  Parametrization of the weights & Based on a relevance function that binarizes the imbalance problem: automatic or defined by the user & Naturally based on the inverse of the kernel density estimate of $y$ or defined by the user \\  
\end{tabular}
\label{table2}
\end{center}
\end{table}

\paragraph{Remark on the definition of the weights:} note that the automatic relevance function with the UBL package does not work for every dataset (error message). We are often asked to define the relevance function which can be difficult for the user. The GOLIATH is pretty reliable because of the simplicity of the approach. However, it is important to define relevant safeguards for the definition of the weight, especially for the very extreme value which presents a very low probability and so a high value of its inverse. We suggest using  a trimming sequence as a hyperparameter, as often proposed in non-parametric statistics inference.

\paragraph{Remark on the computation time:} Both the UBL package and the GOLIATH algorithm are fast enough to generate a new sample: between 3 and 5 seconds for a dataset with about 500 rows. Note that, with the Non-Classical Smoothed Bootstrap, the estimation of the bandwidth parameter for a non-Gaussian distribution could take several minutes due to the package used, especially for a Binomial one.

\subsection{GOLIATH Algorithm}

\subsubsection{GOLIATH Overall Algorithm}
The GOLIATH algorithm is presented in Figure \ref{algoGOLIATH}.
\paragraph{Remarks on the use of GOLIATH:}
\bit
\item As described in the paper, GOLIATH proposes 3 modes of data generation: "synth" to obtain a complete synthetic dataset, "augment" to obtain the original dataset augmented with synthetic observations, and "mix" which is a mixed approach: the original sample for the first occurrence of the seed drawn and a synthetic observation for the next.
\item The automatic weights used are defined as the inverse of the kernel density estimate of the $y$ distribution with the mod "mix" and "synth" because a new dataset is built which is equivalent to realizing an oversampling and an undersampling. However, with the mod "augment", the weights are defined as the squared inverse $\frac{1}{\widehat{f}^2}$ because we realize an oversampling and want to draw more extreme values.
\item The parameters of GOLIATH depend on the generator used: the number of the nearest neighbors $k$, the tuning noise for Classical and Non-Classical Smoothed Bootstrap equivalent to $hmult$ for the ROSE algorithm, and $pert$ for the Gaussian Noise algorithm.
\item It is possible to define another distance with the Nearest Neighbors Smoothed Boostrap. GOLIATH computes the $k$-NN using a distance in the R-package \textit{philentropy} that proposes a large choice of distance.  
\item It is possible to use GOLIATH as a simple generator of data (non-supervised framework) with a $Y$ defined as null. This corresponds to the first step of GOLIATH. It is also possible to use GOLIATH in a classical (non-imbalanced) supervised framework (with a $Y$ non-null) to perform the learning and prediction.
\item The clustering is an option. If activated, it is possible to define a clustering based on the initial train dataset. It is also possible to define a clustering on $Y$ distribution in order to define the clusters according to the frequencies of $Y$ values, in the same  idea as the UBL approach.  
\item The $m$ parameter represents the maximum number of components in the Gaussian Mixture Model used for the clustering. Thus, this algorithm seeks to optimize the clustering with a number of clusters between 1 and $m$. Obviously, it is possible to use another clustering algorithm as k-means.
\eit

\subsubsection{Important Required Packages}

The GOLIATH algorithm uses the following R-packages:
\bit
\item \textit{Ake: Associated Kernel Estimations, used for the bandwidth parameter of the Binomial distributions} (\url{https://cran.r-project.org/web/packages/Ake/Ake.pdf})
\item \textit{ks: Kernel Smoothing,  used for the bandwidth parameter of the Gaussian distributions} \url{https://cran.r-project.org/web/packages/ks/ks.pdf}
\item \textit{np: Nonparametric Kernel Smoothing Methods for Mixed Data Types,  used for the bandwidth parameter of the Beta, truncated Gaussian and Gamma distributions}  \url{https://cran.r-project.org/web/packages/np/np.pdf}
\item \textit{kernelboot: Smoothed Bootstrap and Random Generation from Kernel Densities, used for the classical smoothed bootstrap} \url{https://cran.r-project.org/web/packages/kernelboot/kernelboot.pdf}
\item \textit{randomForest: Breiman and Cutler's Random Forests for Classification and Regression, used for the estimation of $y$ for the new synthetic data} \url{https://cran.r-project.org/web/packages/randomForest/randomForest.pdf}
\item \textit{mclust: Gaussian Mixture Modelling for Model-Based Clustering, Classification, and Density Estimation, used for the GMM clustering} (\url{https://cran.r-project.org/web/packages/mclust/mclust.pdf})
\item \textit{philentropy: Similarity and Distance Quantification Between Probability Functions, used for the computation of the $k-$ nearest neighbors in the Nearest Neighbors Smoothed Boostrap} (\url{https://cran.r-project.org/web/packages/philentropy/philentropy.pdf})
\eit

\subsubsection{GOLIATH algorithm for Y}

We proposed to generate a synthetic target value  $y^*$ associated to a new synthetic covariate $x^*$ using the wild bootstrap approach. This technique is well-known in the regression framework, especially in the presence of heteroskedasticity.

\begin{algorithm}[H]
   \caption{GOLIATH algorithm for Y}
   \label{algoY}
\begin{algorithmic}
    \scriptsize
    \STATE {\bfseries Input} \textit{covariates} X; \textit{target variable} Y; \textit{synthetic X} synth, \textit{drawn samples for the synthetic X} seed; method for the generation of $y$ method\_Y=1; \textit{standard deviation for the Gaussian distribution of the noise in the Wild Bootstrap} sigma=0, \textit{proportion of the perturbation for the Gaussian Noise method} pert\\
    {\bf RF prediction}  \hskip3em// \textit{Prediction with the Random Forest algorithm}\\
    \STATE model = RF(X,Y) \hskip3em// \textit{Training the model}\\
    \STATE predSynth = predict(model, synth)\hskip3em// \textit{Prediction on the synthetic data}\\
    \STATE predSeed = predict(model, X[seed])\hskip3em// \textit{Prediction on the seed data}\\
    \STATE eps = Y[seed] - predSeed \hskip3em// \textit{Distribution of the prediction error}\\
    {\bf Y Generation}  \hskip3em// \textit{generation of synthetic Y using the prediction error}\\
    \STATE For i in synth
        \STATE \hskip1em  if method\_Y = 1 then \\ 
            \STATE \hskip2em  eps = Y[seed][i] - DistribPredSeed[i]  \hskip3em// \textit{Distribution of the absolute residuals on the prediction of the seed}\\
            \STATE \hskip2em v = Gaussian(0,sigma) \hskip3em// \textit{Noise in the wild bootstrap}\\
            \STATE \hskip2em  synth[i,Y]=Y[seed][i] + abs(Random(eps,1)) * v * sign(AveragePredSeed[i]-AveragePredSynth[i]) \hskip2cm
        \STATE \hskip1em else if method\_Y = 0 then \\ 
            \STATE \hskip2em  synth[i,Y]=Y[seed][i] \\
        \STATE \hskip1em else if method\_Y = 2 then \\ 
            \STATE \hskip2em  eps = AveragePredSynth[i] - AveragePredSeed[i]  \hskip3em// \textit{difference between the average prediction}\\
            \STATE \hskip2em  synth[i,Y]=Y[seed][i] + eps \\
        \STATE \hskip1em else if method\_Y = 3 then \\ 
            \STATE \hskip2em  eps = Y[seed][i] - DistribPredSeed[i]  \hskip3em// \textit{Distribution of the absolute residuals on the prediction of the seed}\\
            \STATE \hskip2em v = Gaussian(0,sigma) \hskip3em// \textit{Noise in the wild bootstrap}\\
            \STATE \hskip2em  h = kde(eps)  \hskip3em// \textit{bandwidth parameter with a gaussian kernel}\\
            \STATE \hskip2em  synth[i,Y]=Y[seed][i] + abs(Gaussian(0,h)) * v * sign(AveragePredSeed[i]-AveragePredSynth[i]) \\
        \STATE \hskip1em else if method\_Y = 4 then \\ 
            \STATE \hskip2em  sig = standardDeviation(Y,w\_Y)  \hskip3em// \textit{weighted standard deviation of $y$}\\
            \STATE \hskip2em  synth[i,Y]=Y[seed][i] + abs(Gaussian(0,sig * pert)) * sign(AveragePredSeed[i]-AveragePredSynth[i]) \\
        \STATE \hskip1em else if method\_Y = 5 then \\ 
            \STATE \hskip2em  h = kde(Y,w\_Y)  \hskip3em// \textit{bandwidth parameter with a gaussian kernel for the weighted kernel density estimate of $y$}\\
            \STATE \hskip2em  synth[i,Y]=Y[seed][i] + abs(Gaussian(0,h)) * sign(AveragePredSeed[i]-AveragePredSynth[i]) \\
    \STATE end For \\
    return Synth[,Y]
\end{algorithmic}
\end{algorithm}

\paragraph{Details of the methods:}
\bit
\item \textbf{method 0:} the synthetic $y^*$ is equal to the $y$ seed i.e. the value of $y$ associated to the $X$ used for generating the new synthetic $x^*$
\item \textbf{method 1 (by default):} the synthetic $y^*$ is generated by using an adapted wild bootstrap technique, as described in the paper. Note that with a sigma parameter equal to 0, we obtain the residual resampling technique: this is the technique used by default.
\item \textbf{method 2:} the synthetic $y^*$ is generated by adding to the $y$ seed the difference between the seed prediction and the synthetic prediction. The idea is to shift the $y$ seed with the difference of the predictions representing the impact of the synthetic $x^*$ in according to the seed $x$ 
\item \textbf{method 3:} the synthetic $y^*$ is generated by using a classical smoothed bootstrap on the distribution of the error prediction. This is a "smoothing" version of the method 1. 
\item \textbf{method 4:} the synthetic $y^*$ is generated by using an adapted Gaussian Noise used a weighted standard deviation of $y$ and the sign of the difference between the predictions of $y$ and $y^*$ 
\item \textbf{method 5:} the synthetic $y^*$ is generated by using a classical smoothed bootstrap and the sign of the difference between the predictions of $y$ and $y^*$.
\eit

As detailed in the paper, method 1 for the generation of $y$  proposes to use an adapted wild bootstrap. The differences with a classical wild bootstrap are:
\bit
\item GOLIATH draw belongs to the distribution of the error prediction for the same $y$ while the Wild Bootstrap draw belongs to the distribution of the error prediction on the whole training dataset.
\item GOLIATH generates a $y$ value for a new $x^*$ sample while the Wild Bootstrap generates a $y$ value for an existing $x$ drawn in the training dataset.
\item To consider the previous item, GOLIATH suggests using the sign between the average prediction of $y$ associated with $x^*$ and associated with $x$. This represents the impact of generating new synthetic data $x^*$ from a seed $x$.
\eit

\begin{figure}[ht]
\centering
\begin{subfigure}{0.49\textwidth}
    \includegraphics[width=1\textwidth]{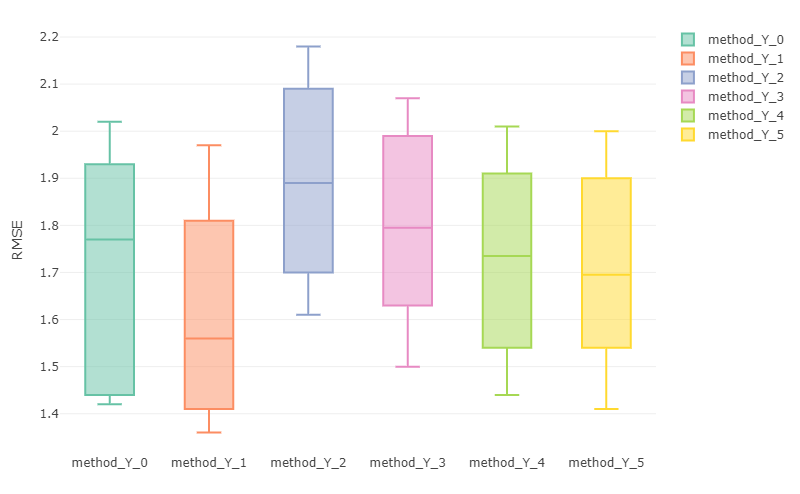}
    \caption{Boxplots of RMSE obtained for different generation methods of $y$, for the same values of synthetic data $x$ obtained with a Classical Smoothed Bootstrap}
    \label{RMSE_methodY}
\end{subfigure}
\hfill
\begin{subfigure}{0.49\textwidth}
    \includegraphics[width=1\textwidth]{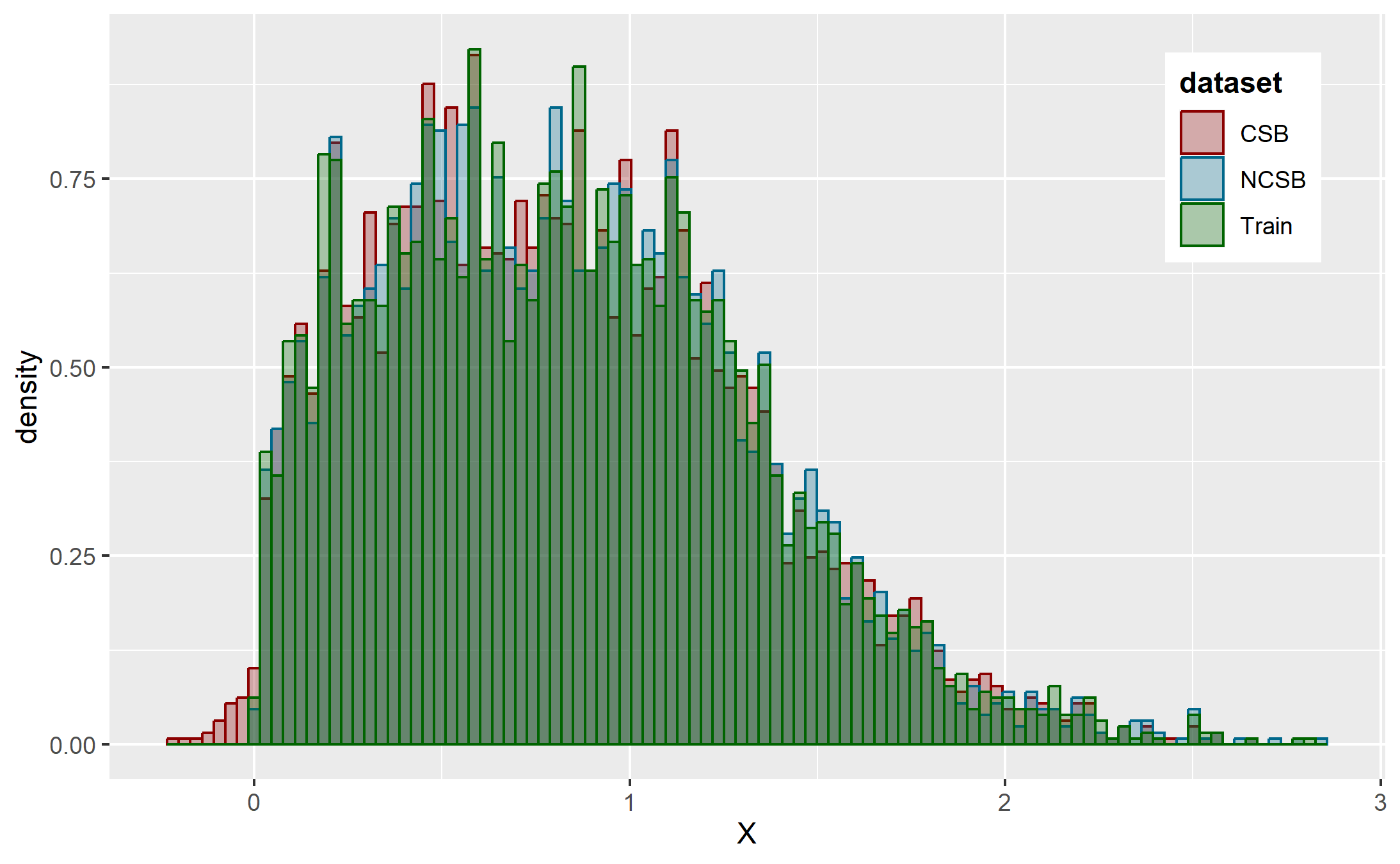}
    \caption{Illustration of a Classical Smoothed Bootstrap vs Non-Classical Smoothed Bootstrap}
    \label{comp_X_CSB-vs-NCSB}
\end{subfigure}
\caption{Illustrations of GOLIATH}
\end{figure}

Figure \ref{RMSE_methodY} shows the difference in the RMSE with the different methods. These results are obtained on the illustrative application with a Classical Smoothed Bootstrap.

\subsubsection{Illustration of the generators}
Figure \ref{comp_X_CSB-vs-NCSB} represents the distributions of the positive variable \textit{WholeWeight} in the dataset \textit{Abalone}. As we can see, the Classical Smoothed Bootstrap can generate synthetic data beyond the $X$ support. Indeed, a Classical Smoothed Bootstrap is based on a classical kernel which is symmetric distribution. For a given point (0 for example) there is as much chance to generate a smaller value (negative) as a larger one (positive). For a positive asymmetric distribution, this can lead to obtaining outliers: as negative values. We observe that the Non-Classical Smoothed Bootstrap, using a Gamma Kernel, generates proper values of $x$.

\subsection{Complementary Results for Illustrative Application}

\paragraph{Remark:} A decomposition by model of the stacked RMSE is not possible because the 10 models are not the same for each run. 

\subsubsection{Dataset details}

\begin{figure}[ht]
\centering
\begin{subfigure}{0.49\textwidth}
    \includegraphics[width=1\textwidth]{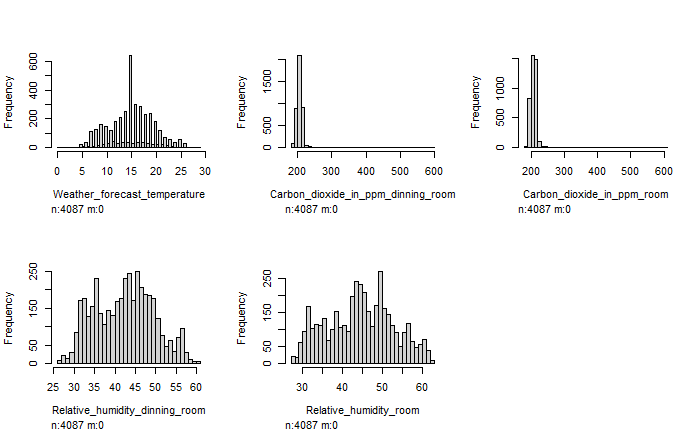}
    \caption{Histograms of the covariates $x$ in the original dataset}
\end{subfigure}
\hfill
\begin{subfigure}{0.49\textwidth}
    \includegraphics[width=1\textwidth]{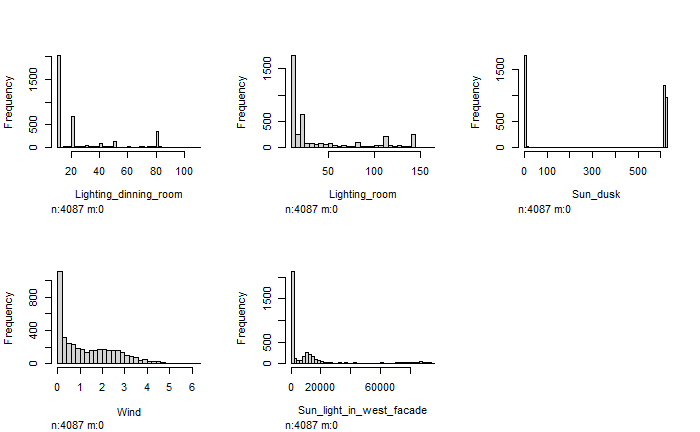}
    \caption{Histograms of the covariates $x$ in the original dataset}
\end{subfigure}
\begin{subfigure}{0.49\textwidth}
    \includegraphics[width=1\textwidth]{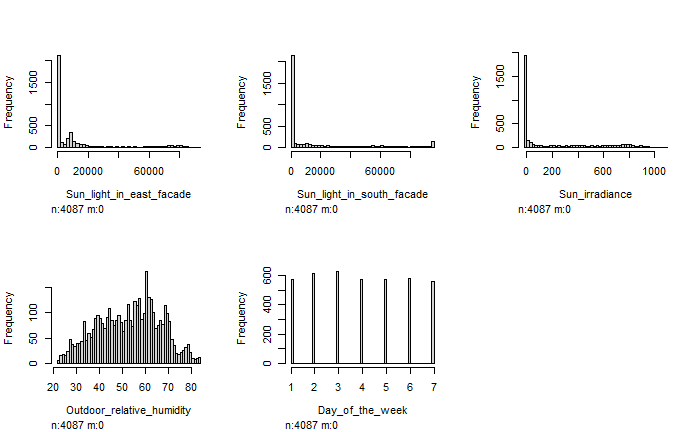}
    \caption{Histograms of the covariates $x$ in the original dataset}
\end{subfigure}
\hfill
\begin{subfigure}{0.49\textwidth}
    \includegraphics[width=1\textwidth]{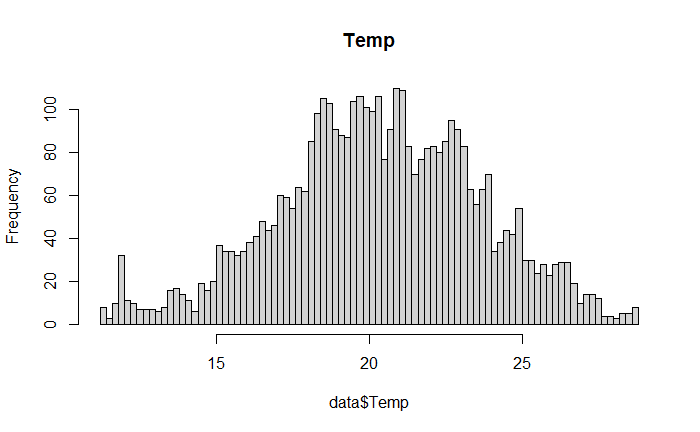}
    \caption{Histograms of the target variable $Y$ in the original dataset}
\end{subfigure}
\caption{Histograms of $X$ and $y$ in the illustrative application dataset}
\label{Histogram_Illu}
\end{figure}

The dataset used in the illustrative application is  \textit{SML2010 Data Set} from the Machine Learning repository UCI (\url{https://archive.ics.uci.edu/ml/datasets/SML2010}). It is composed of 24 numeric attributes and 4137 instances. The target variable is the indoor temperature (we construct a unique target variable as the mean temperature of dinning-room and the temperature of the room). Figure \ref{Histogram_Illu} gives the histograms for all covariates $X$ and the target variable $y$. It can be observed that some variables (lighting and wind for instance) are bounded because of their positivity. 

\begin{table}[H]
\caption{Descriptive statistics of the dataset} 
\scriptsize
\begin{center}
\begin{tabular}{ | c|| c| c || c || c || c || c | } 
Variable & Min. & 1st Qu. & Median & Mean & 3rd Qu. & Max. \\
\hline
Weather\_forecast\_temperature & 0 & 12 & 15 & 15.1 & 18 & 29\\
\hline
Carbon\_dioxide\_in\_ppm\_dinning\_room & 187.34 & 200.33 & 205.22 & 206.72 & 210.07 & 594.39\\
\hline
Carbon\_dioxide\_in\_ppm\_room & 188.91 & 201.83 & 208.97 & 209.74 & 212.37 & 609.24\\
\hline
Relative\_humidity\_dinning\_room & 26.17 & 36.02 & 42.73 & 42.31 & 47.5 & 60.96\\
\hline
Relative\_humidity\_room & 27.26 & 38.33 & 44.71 & 44.46 & 50.2 & 62.59\\
\hline
Lighting\_dinning\_room & 10.74 & 11.56 & 14.33 & 29.11 & 40.75 & 111.8\\
\hline
Lighting\_room & 11.33 & 13.51 & 22.21 & 42.56 & 55.28 & 162.96\\
\hline
Rain & 0 & 0 & 0 & 0.03 & 0 & 1\\
\hline
Sun\_dusk & 0.61 & 0.65 & 612.95 & 335.72 & 619.76 & 625\\
\hline
Wind & 0 & 0.17 & 0.96 & 1.3 & 2.23 & 6.32\\
\hline
Sun\_light\_in\_west\_facade & 0 & 0 & 831.49 & 14876.48 & 14691.85 & 95278.4\\
\hline
Sun\_light\_in\_east\_facade & 0 & 0 & 1125.38 & 13680.87 & 13108.25 & 92367.5\\
\hline
Sun\_light\_in\_south\_facade & 0 & 0 & 716.8 & 20028.33 & 34069.8 & 95704.4\\
\hline
Sun\_irradiance & -4.16 & -3.25 & 12.22 & 234.14 & 488.37 & 1094.66\\
\hline
Outdoor\_relative\_humidity & 22.25 & 42.46 & 54.38 & 53.07 & 62.89 & 83.81\\
\hline
Day\_of\_the\_week & 1 & 2 & 4 & 3.96 & 6 & 7\\
\hline
Temp & 11.21 & 18.26 & 20.31 & 20.33 & 22.66 & 28.73\\
\hline
\end{tabular}
\label{table3}
\end{center}
\end{table}

\subsubsection{Protocol}

\begin{figure}[H]
\centering
\begin{subfigure}{0.49\textwidth}
    \includegraphics[width=1\textwidth]{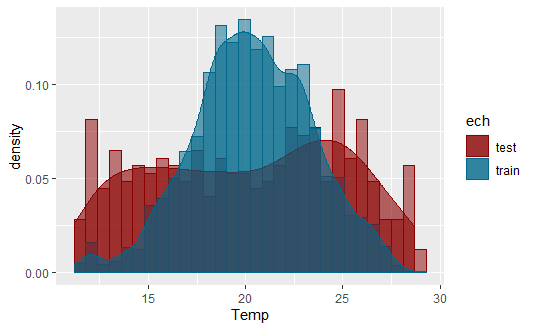}
    \caption{Example of a train-test set}
    \label{trainTest}
\end{subfigure}
\hfill
\begin{subfigure}{0.49\textwidth}
    \includegraphics[width=1\textwidth]{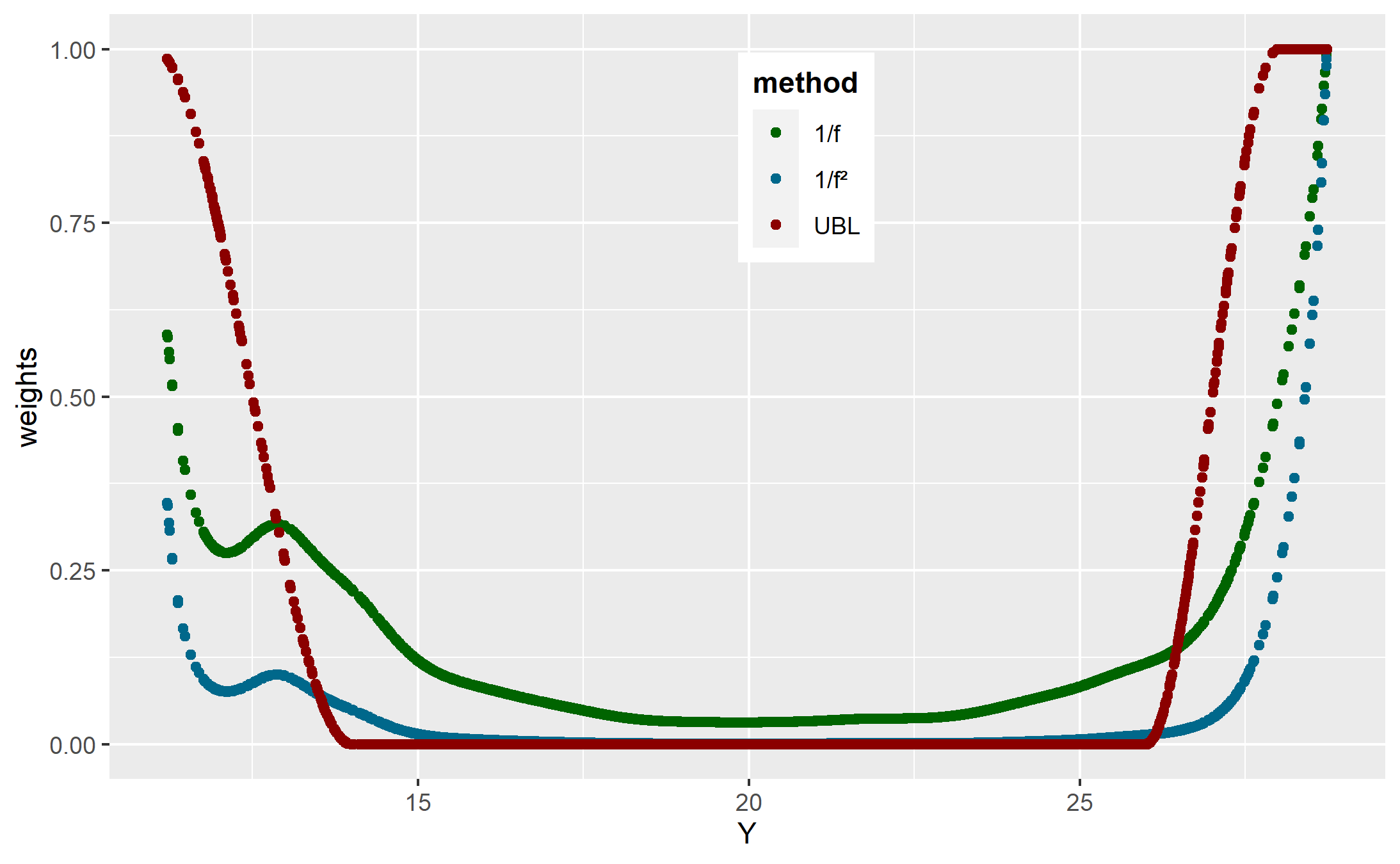}
    \caption{Comparison of the weighting method}
    \label{Weighting}
\end{subfigure}
\caption{Illustration of the protocol}
    \label{protocol}
\end{figure}

The protocol for the experiments on the illustrative application can be summarized as follows:
\begin{enumerate}
    \item Define $test\_prop$ the desired proportion of the test dataset: 10\% here
    \item Define $N\_sample$ the number of the runs i.e. the desired train-test set: 10 here
    \item Define the proportion of the imbalanced dataset $imb\_prop$: 10\% here.
    \item Construct $N\_sample$ train-test set: repeat the following instructions:
    \bit
        \item draw a test sample with a size $size(data) \times test\_prop$,
        \item draw $size(data-test) \times imb\_prop$ from the remaining dataset with weights squared in order to get slightly more rare observations: on the side here. An illustration is given on Figure \ref{trainTest}.    
    \eit
    \item Generate the new train datasets with the different methods. The generation is based on a weighting function. In Figure \ref{Weighting}, we compare 
    \bit
        \item the weights obtained with the inverse of the kernel density estimate of the $y$ distributions: giving a distribution $y$ approximately uniform, 
        \item the weights obtained with the squared inverse of the kernel density estimate of the $y$ distributions: giving a distribution $y$ inverse to that of $y$ in the initial sample,
        \item the weights obtained with the UBL approach.
    \eit
    \item Predict the test value according to the new train datasets
\end{enumerate}

\subsubsection{Predictive performance metrics}

\begin{figure}[ht]
\centering
\begin{subfigure}{0.49\textwidth}
    \includegraphics[width=1\textwidth]{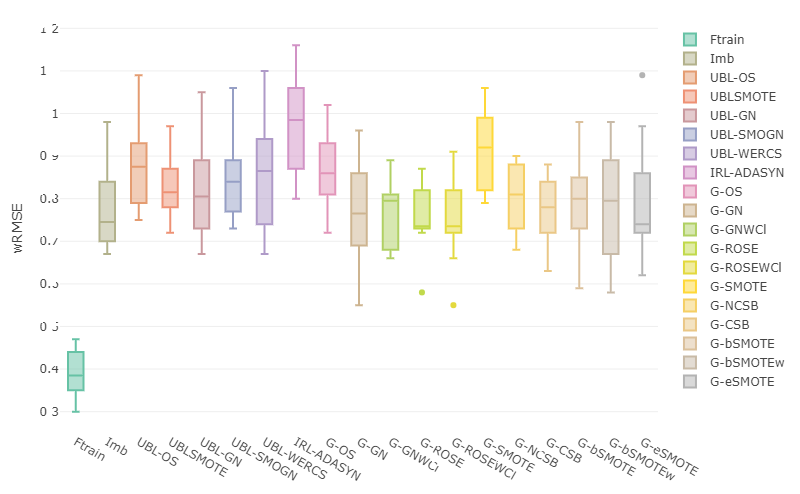}
    \caption{Weighted RMSE Boxplots}
\end{subfigure}
\hfill
\begin{subfigure}{0.49\textwidth}
    \includegraphics[width=1\textwidth]{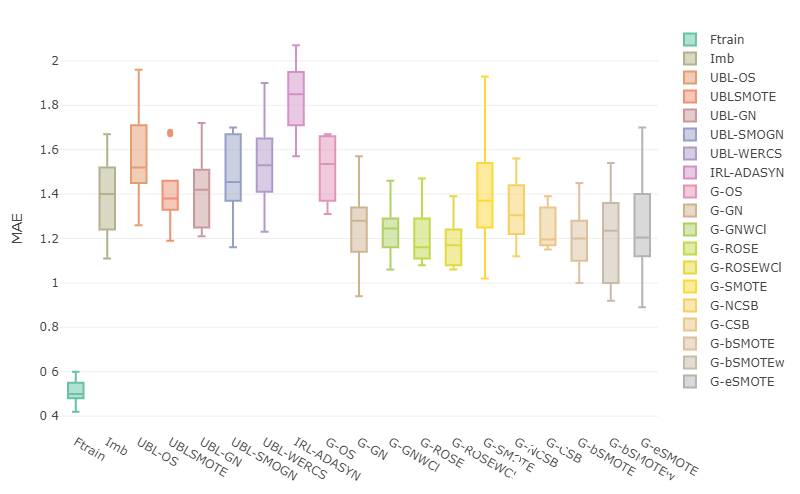}
    \caption{MAE boxplots}
\end{subfigure}
\caption{Boxplots of weighted RMSE and MAE for 10 runs}
    \label{wRMSEMAE_AppliIllu}
\end{figure}

The results for the MAE are similar to the RMSE results. The weighted-RMSE is still better with GOLIATH algorithm than UBL approach but GOLIATH-SMOTE presents an important value. The weighted-RMSE is clearly better with the (extended-) Nearest Neighbors Smoothed Bootstrap.

\subsubsection{Results for 20 models}

\begin{figure}[H]
\centering
\begin{subfigure}{0.32\textwidth}
    \includegraphics[width=\textwidth]{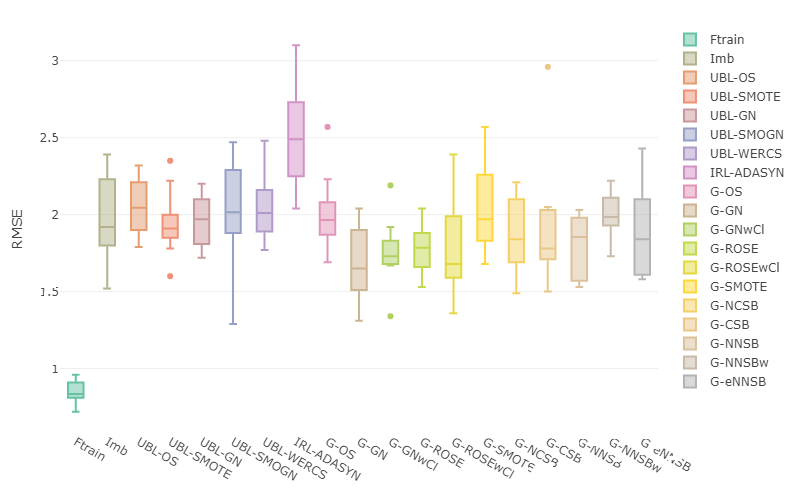}
    \caption{RMSE Boxplots for 20 models}
\end{subfigure}
\begin{subfigure}{0.32\textwidth}
    \includegraphics[width=\textwidth]{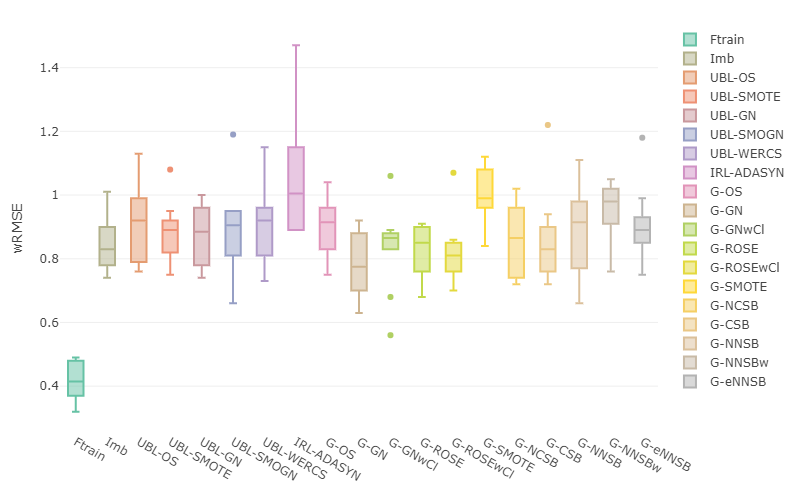}
    \caption{weighted-RMSE Boxplots for 20 models}
\end{subfigure}
\begin{subfigure}{0.32\textwidth}
    \includegraphics[width=\textwidth]{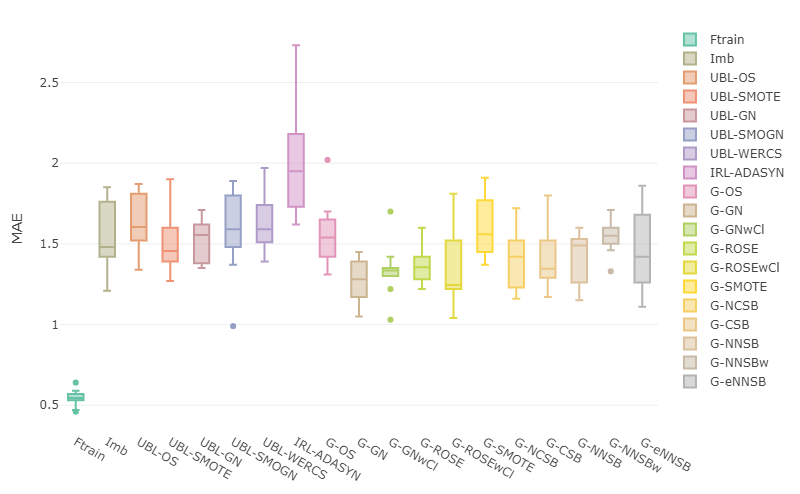}
    \caption{MAE Boxplots for 20 models}
\end{subfigure}
\caption{Boxplots of predictive performance metrics for 20 runs}
    \label{Results_20models}
\end{figure}

The results for the 20 models generally confirm those obtained with 10 runs.  However, GOLIATH-SMOTE seems quite worse than the other GOLIATH techniques. It is important to see that the Nearest Neighbors Smoothed Bootstrap improves these results.

\subsubsection{Results for 20 runs}

\begin{figure}[H]
\centering
\begin{subfigure}{0.32\textwidth}
    \includegraphics[width=\textwidth]{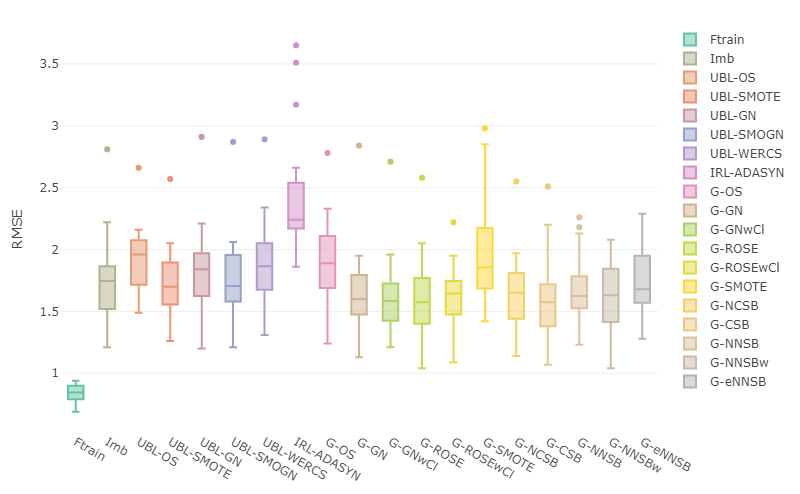}
    \caption{RMSE Boxplots for 20 runs}
\end{subfigure}
\begin{subfigure}{0.32\textwidth}
    \includegraphics[width=\textwidth]{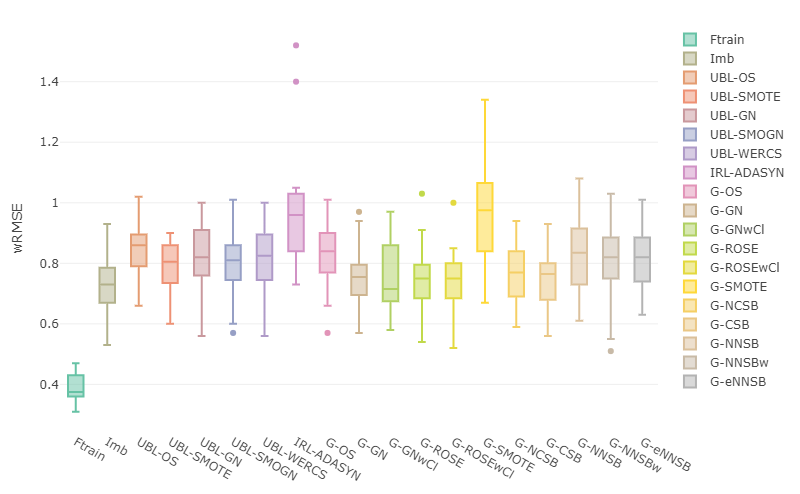}
    \caption{weighted-RMSE Boxplots for 20 runs}
\end{subfigure}
\begin{subfigure}{0.32\textwidth}
    \includegraphics[width=\textwidth]{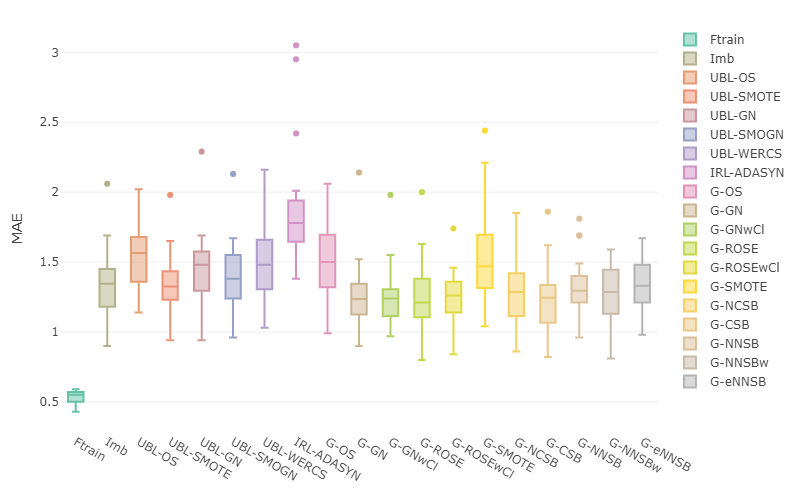}
    \caption{MAE Boxplots for 20 runs}
\end{subfigure}
\caption{Boxplots of predictive performance metrics for 20 runs}
    \label{Results_20runs}
\end{figure}

The results for the 20 runs confirm those obtained with 10 runs. 

\subsubsection{IRon specific metrics for Imbalanced Regression}

The R-package \textit{IRon: Solving Imbalanced Regression Tasks} (\url{https://cran.r-project.org/web/packages/IRon/IRon.pdf}), is a useful and relevant package specific to Imbalanced Regression. It is based on \cite{ribeiro2020imbalanced} and offers several adapted metrics.
Below, we propose an analysis of these predictive performance metrics in order to compare the approaches.  

\begin{figure}[H]
\centering
\begin{subfigure}{0.32\textwidth}
    \includegraphics[width=\textwidth]{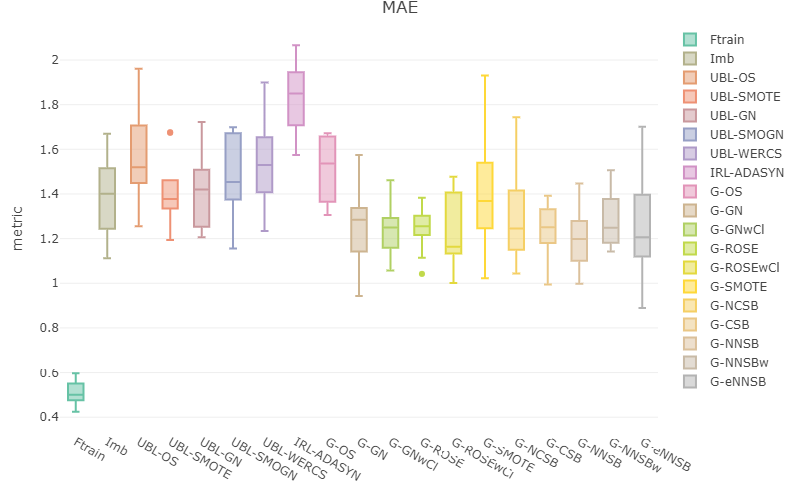}
    \caption{MAE Boxplots}
\end{subfigure}
\begin{subfigure}{0.32\textwidth}
    \includegraphics[width=\textwidth]{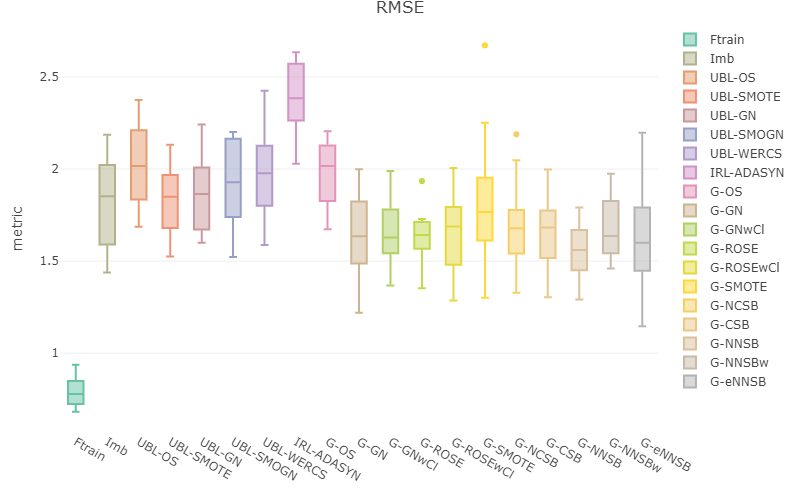}
    \caption{RMSE Bowplots}
\end{subfigure}
\begin{subfigure}{0.32\textwidth}
    \includegraphics[width=\textwidth]{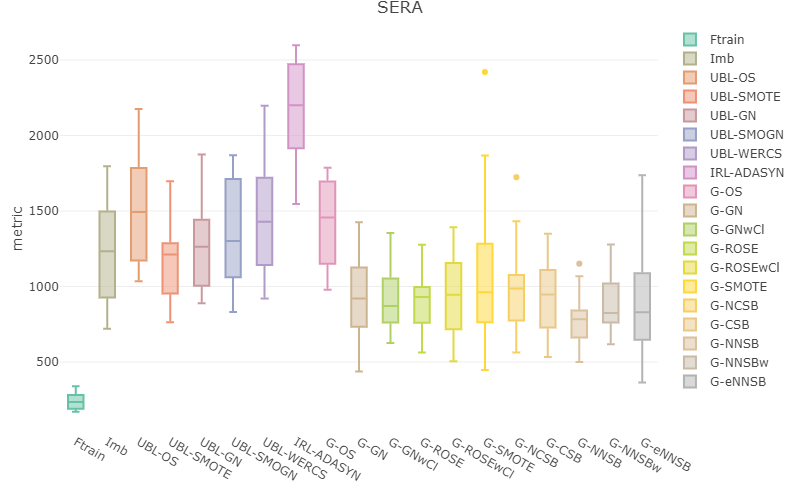}
    \caption{SERA Boxplots}
\end{subfigure}
\begin{subfigure}{0.32\textwidth}
    \includegraphics[width=\textwidth]{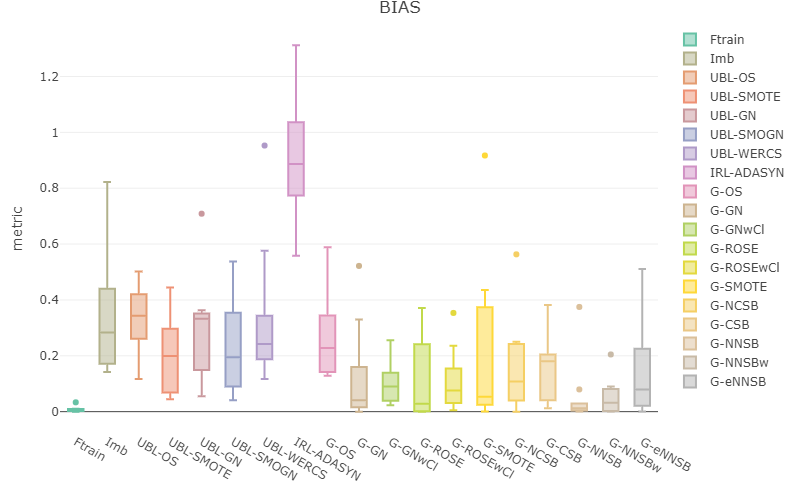}
    \caption{BIAS Boxplots}
\end{subfigure}
\begin{subfigure}{0.32\textwidth}
    \includegraphics[width=\textwidth]{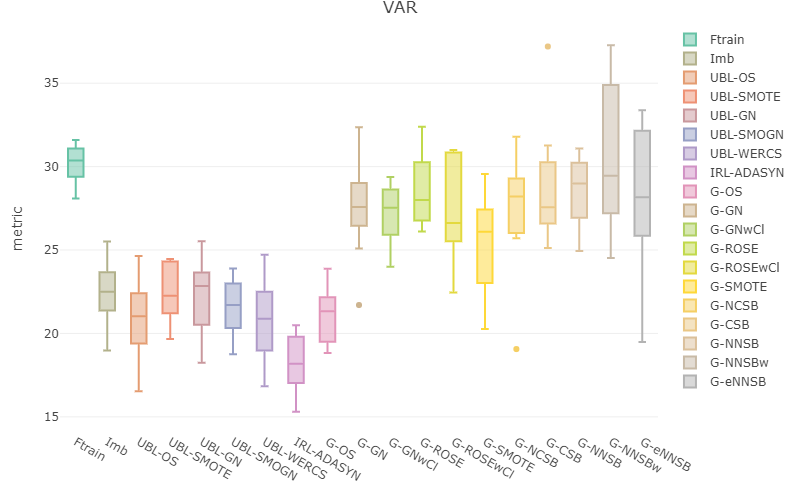}
    \caption{VAR Boxplots}
\end{subfigure}
\caption{Boxplots of predictive performance metrics with IRon package}
    \label{IRon_metrics}
\end{figure}

The results obtained with the IRon package confirm those obtained from our metrics: The RMSE, MAE, and SERA present the same look. The Biases are lower for GOLIATH but the variances are higher.

\subsection{Complementary Results for Imbalanced Regression Applications}

\subsubsection{Protocol}

The protocol for the applications is quite similar to the previous one for the illustrative application. It can be summarized as follow:
\begin{enumerate}
    \item Data preprocessing: removal of eventual categorical covariates, eventual conversion of some covariates, removal of missing data.
    \item Define $test\_prop$ the desired proportion of the test dataset:  cf below.
    \item Define $N\_sample$ the number of the runs i.e. the desired train-test set: 10 here.
    \item Define an eventual proportion of the imbalanced dataset $imb\_prop$ to obtain an extremely imbalanced dataset: cf below.
    \item Construct $N\_sample$ train-test set: repeat the following instructions:
    \bit
        \item draw a test sample with a size $size(data) \times test\_prop$,
        \item draw $size(data-test) \times imb\_prop$ from the remaining dataset with weights squared.
    \eit
    \item Generate the new train datasets with the different methods. The generation is based on a weighting function. 
    \item Predict the test value according to the new train datasets.
\end{enumerate}

$test\_prop$: Abalone: 30\%, Bank8FM: 50\%, Boston: 30\%, CpuSm: 5\%, NO2: 10\% \\
$imb\_prop$: Abalone: 10\%, Bank8FM: 10\%, Boston: 100\%, CpuSm: 5\%, NO2: 100\%

\subsubsection{Details for the Abalone dataset}

The Abalone dataset is composed of 4177 observations and 8 numerical variables including 0 discrete. More details on the covariates and the target variable are given below. We can observe especially on the histograms the distribution and the eventual boundaries of the variables. 

\begin{figure}[H]
\centering
\begin{subfigure}{0.32\textwidth}
    \includegraphics[width=\textwidth]{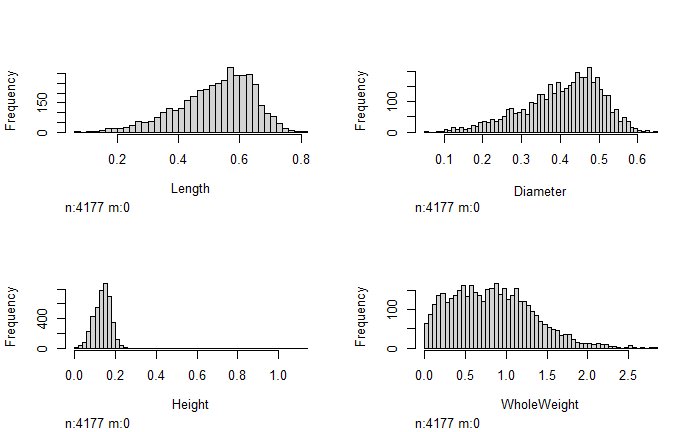}
    \caption{Histograms of the covariates $x$ in the original dataset}
\end{subfigure}
\begin{subfigure}{0.32\textwidth}
    \includegraphics[width=\textwidth]{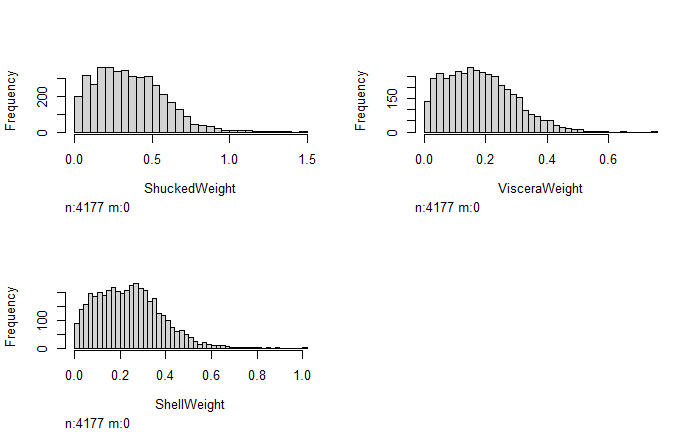}
    \caption{Histograms of the covariates $x$ in the original dataset}
\end{subfigure}
\begin{subfigure}{0.32\textwidth}
    \includegraphics[width=\textwidth]{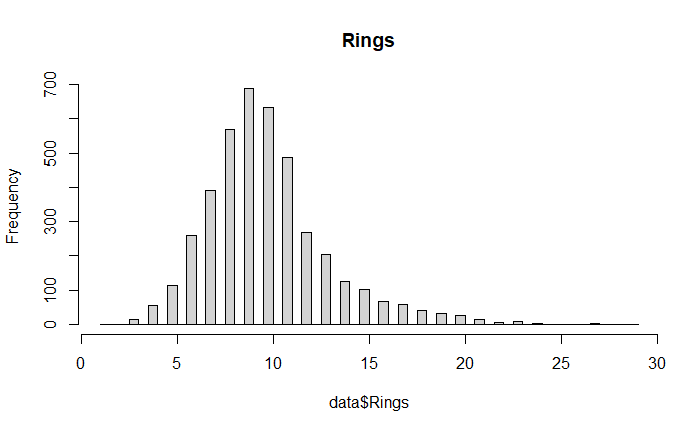}
    \caption{Histogram of the target variable $Y$ in the original dataset}
\end{subfigure}
\caption{Histograms of $X$ and $y$ for the Abalone dataset}
\end{figure}

\begin{table}[H]
\caption{Descriptive statistics of the dataset} 
\scriptsize
\begin{center}
\begin{tabular}{ | c|| c| c || c || c || c || c | } 
Variable & Min. & 1st Qu. & Median & Mean & 3rd Qu. & Max. \\
\hline
Rings & 1 & 8 & 9 & 9.93 & 11 & 29\\
Length & 0.07 & 0.45 & 0.54 & 0.52 & 0.62 & 0.81\\
Diameter & 0.06 & 0.35 & 0.42 & 0.41 & 0.48 & 0.65\\
Height & 0 & 0.12 & 0.14 & 0.14 & 0.16 & 1.13\\
WholeWeight & 0 & 0.44 & 0.8 & 0.83 & 1.15 & 2.83\\
ShuckedWeight & 0 & 0.19 & 0.34 & 0.36 & 0.5 & 1.49\\
VisceraWeight & 0 & 0.09 & 0.17 & 0.18 & 0.25 & 0.76\\
ShellWeight & 0 & 0.13 & 0.23 & 0.24 & 0.33 & 1\\
\hline
\end{tabular}
\label{table4}
\end{center}
\end{table}

\subsubsection{Details for the Bank8FM dataset}

The Bank8FM dataset is composed of 4499 observations and 9 numerical variables including 1 discrete. More details on the covariates and the target variable are given below. We can observe especially on the histograms the distribution and the eventual boundaries of the variables. 

\begin{figure}[H]
\centering
\begin{subfigure}{0.32\textwidth}
    \includegraphics[width=\textwidth]{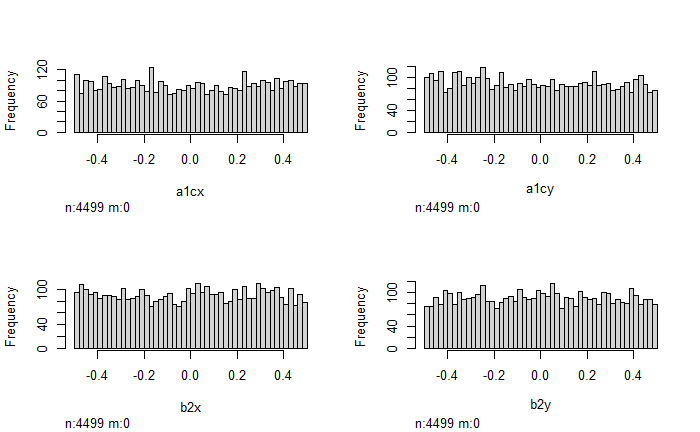}
    \caption{Histograms of the covariates $x$ in the original dataset}
\end{subfigure}
\begin{subfigure}{0.32\textwidth}
    \includegraphics[width=\textwidth]{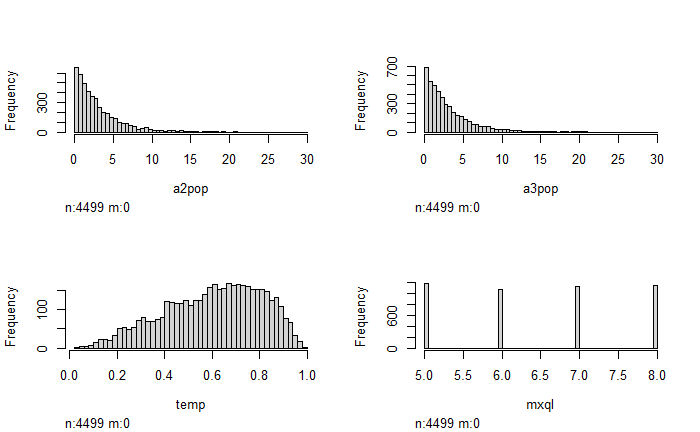}
    \caption{Histograms of the covariates $x$ in the original dataset}
\end{subfigure}
\begin{subfigure}{0.32\textwidth}
    \includegraphics[width=\textwidth]{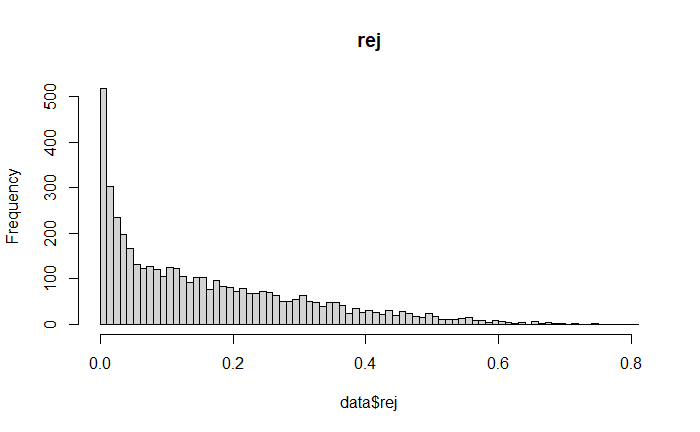}
    \caption{Histogram of the target variable $Y$ in the original dataset}
\end{subfigure}
\caption{Histograms of $X$ and $y$ for the Bank8FM dataset}
\end{figure}

\begin{table}[H]
\caption{Descriptive statistics of the dataset} 
\scriptsize
\begin{center}
\begin{tabular}{ | c|| c| c || c || c || c || c | } 
Variable & Min. & 1st Qu. & Median & Mean & 3rd Qu. & Max. \\
\hline
rej & 0 & 0.03 & 0.12 & 0.16 & 0.25 & 0.8\\
a1cx & -0.5 & -0.26 & 0 & 0 & 0.26 & 0.5\\
a1cy & -0.5 & -0.27 & -0.02 & -0.01 & 0.24 & 0.5\\
b2x & -0.5 & -0.25 & 0.01 & 0 & 0.25 & 0.5\\
b2y & -0.5 & -0.25 & 0 & 0 & 0.25 & 0.5\\
a2pop & 0 & 0.9 & 2.13 & 3.05 & 4.19 & 29.71\\
a3pop & 0 & 0.93 & 2.14 & 3.08 & 4.21 & 29.68\\
temp & 0.04 & 0.46 & 0.63 & 0.6 & 0.77 & 0.98\\
mxql & 5 & 5 & 7 & 6.49 & 8 & 8\\
\hline
\end{tabular}
\label{table5}
\end{center}
\end{table}

\subsubsection{Details for the Boston dataset}

The Boston dataset is composed of 506 observations and 13 numerical variables including 1 discrete. More details on the covariates and the target variable are given below. We can observe especially on the histograms the distribution and the eventual boundaries of the variables.

\begin{figure}[H]
\centering
\begin{subfigure}{0.3\textwidth}
    \includegraphics[width=\textwidth]{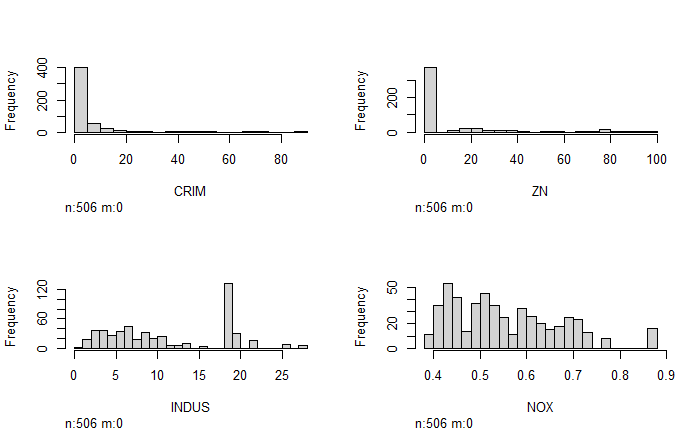}
    \caption{Histograms of the covariates $x$ in the original dataset}
\end{subfigure}
\begin{subfigure}{0.32\textwidth}
    \includegraphics[width=\textwidth]{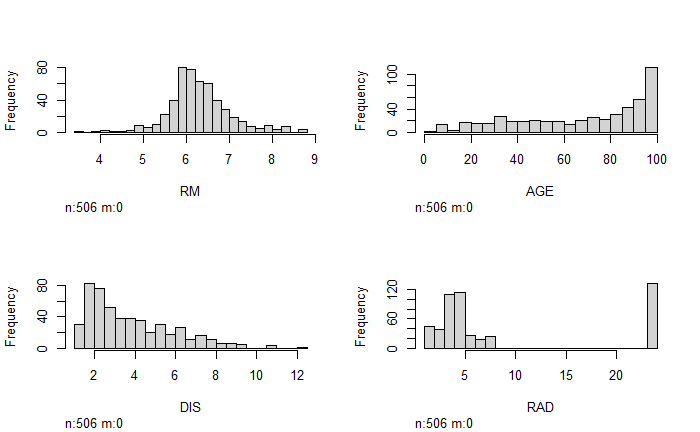}
    \caption{Histograms of the covariates $x$ in the original dataset}
\end{subfigure}
\\
\begin{subfigure}{0.32\textwidth}
    \includegraphics[width=\textwidth]{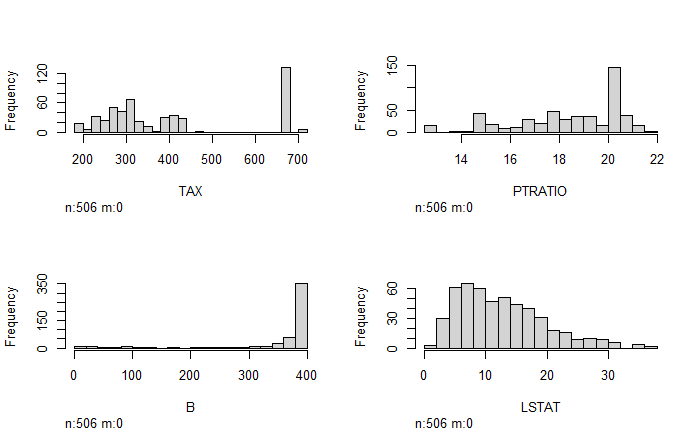}
    \caption{Histograms of the covariates $x$ in the original dataset}
\end{subfigure}
\begin{subfigure}{0.32\textwidth}
    \includegraphics[width=\textwidth]{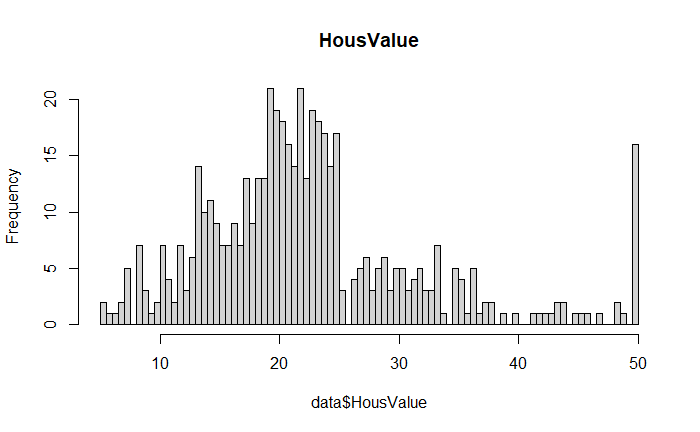}
    \caption{Histogram of the target variable $Y$ in the original dataset}
\end{subfigure}
\caption{Histograms of $X$ and $y$ for the Boston dataset}
\end{figure}

\begin{table}[H]
\caption{Descriptive statistics of the dataset} 
\scriptsize
\begin{center}
\begin{tabular}{ | c|| c| c || c || c || c || c | } 
Variable & Min. & 1st Qu. & Median & Mean & 3rd Qu. & Max. \\
\hline
HousValue & 5 & 17.02 & 21.2 & 22.53 & 25 & 50\\
CRIM & 0.01 & 0.08 & 0.26 & 3.61 & 3.68 & 88.98\\
ZN & 0 & 0 & 0 & 11.36 & 12.5 & 100\\
INDUS & 0.46 & 5.19 & 9.69 & 11.14 & 18.1 & 27.74\\
NOX & 0.38 & 0.45 & 0.54 & 0.55 & 0.62 & 0.87\\
RM & 3.56 & 5.89 & 6.21 & 6.28 & 6.62 & 8.78\\
AGE & 2.9 & 45.02 & 77.5 & 68.57 & 94.07 & 100\\
DIS & 1.13 & 2.1 & 3.21 & 3.79 & 5.19 & 12.13\\
RAD & 1 & 4 & 5 & 9.55 & 24 & 24\\
TAX & 187 & 279 & 330 & 408.24 & 666 & 711\\
PTRATIO & 12.6 & 17.4 & 19.05 & 18.46 & 20.2 & 22\\
B & 0.32 & 375.38 & 391.44 & 356.67 & 396.22 & 396.9\\
LSTAT & 1.73 & 6.95 & 11.36 & 12.65 & 16.96 & 37.97\\

\hline
\end{tabular}
\label{table6}
\end{center}
\end{table}

\subsubsection{Details for the CpuSm dataset}

The CpuSm dataset is composed of 8192 observations and 13 numerical variables including 0 discrete. More details on the covariates and the target variable are given below. We can observe especially on the histograms the distribution and the eventual boundaries of the variables.

\begin{figure}[H]
\centering
\begin{subfigure}{0.32\textwidth}
    \includegraphics[width=\textwidth]{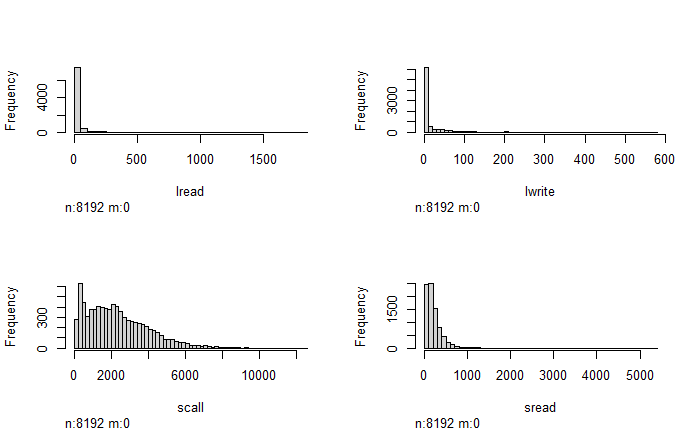}
    \caption{Histograms of the covariates $x$ in the original dataset}
\end{subfigure}
\begin{subfigure}{0.32\textwidth}
    \includegraphics[width=\textwidth]{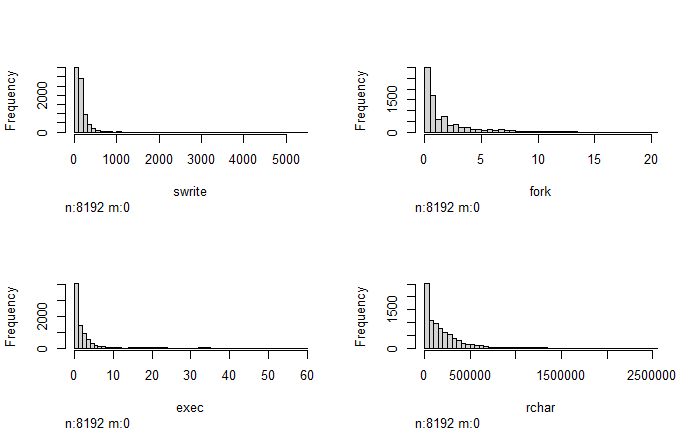}
    \caption{Histograms of the covariates $x$ in the original dataset}
\end{subfigure}
\\
\begin{subfigure}{0.32\textwidth}
    \includegraphics[width=\textwidth]{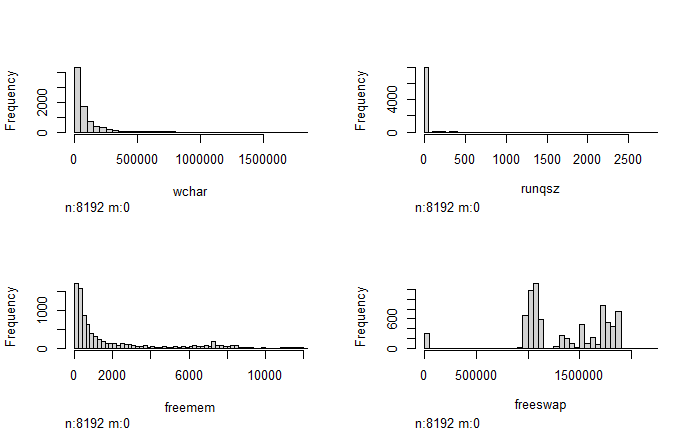}
    \caption{Histograms of the covariates $x$ in the original dataset}
\end{subfigure}
\begin{subfigure}{0.32\textwidth}
    \includegraphics[width=\textwidth]{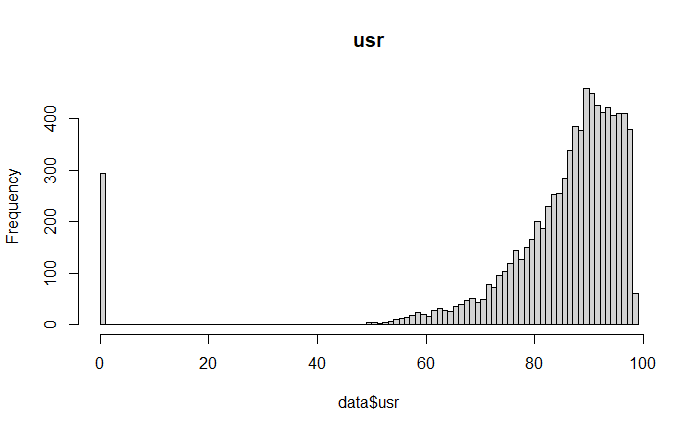}
    \caption{Histogram of the target variable $Y$ in the original dataset}
\end{subfigure}
\caption{Histograms of $X$ and $y$ for the CpuSm dataset}
\end{figure}

\begin{table}[H]
\caption{Descriptive statistics of the dataset} 
\scriptsize
\begin{center}
\begin{tabular}{ | c|| c| c || c || c || c || c | } 
Variable & Min. & 1st Qu. & Median & Mean & 3rd Qu. & Max. \\
\hline
usr & 0 & 81 & 89 & 83.97 & 94 & 99 \\
lread & 0 & 2 & 7 & 19.56 & 20 & 1845 \\
lwrite & 0 & 0 & 1 & 13.11 & 10 & 575 \\
scall & 109 & 1012 & 2051.5 & 2306.32 & 3317.25 & 12493 \\
sread & 6 & 86 & 166 & 210.48 & 279 & 5318 \\
swrite & 7 & 63 & 117 & 150.06 & 185 & 5456 \\
fork & 0 & 0.4 & 0.8 & 1.88 & 2.2 & 20.12 \\
exec & 0 & 0.2 & 1.2 & 2.79 & 2.8 & 59.56 \\
rchar & 278 & 33864.25 & 124779.5 & 197013.67 & 267669.25 & 2526649 \\
wchar & 1498 & 22935.5 & 46620 & 95898.29 & 106148 & 1801623 \\
runqsz & 1 & 1.2 & 2 & 19.63 & 3 & 2823 \\
freemem & 55 & 231 & 579 & 1763.46 & 2002.25 & 12027 \\
freeswap & 2 & 1042623.5 & 1289289.5 & 1328125.96 & 1730379.5 & 2243187 \\
\hline
\end{tabular} 
\label{table7}
\end{center}
\end{table}

\subsubsection{Details for the NO2 dataset}

The NO2 dataset is composed of 500 observations and 8 numerical variables including 0 discrete. More details on the covariates and the target variable are given below. We can observe especially on the histograms the distribution and the eventual boundaries of the variables. 

\begin{figure}[H]
\centering
\begin{subfigure}{0.32\textwidth}
    \includegraphics[width=\textwidth]{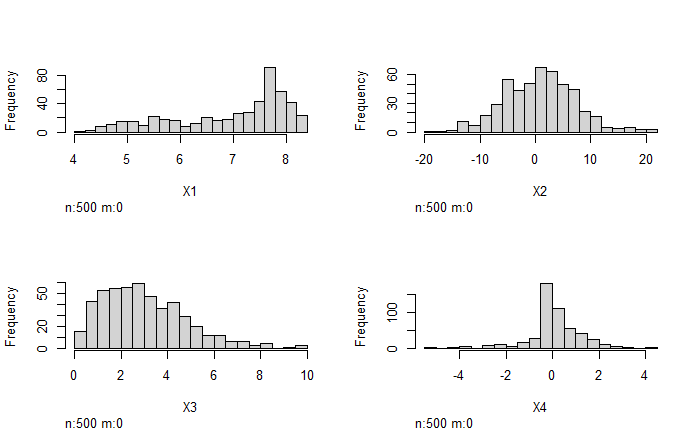}
    \caption{Histograms of the covariates $x$ in the original dataset}
\end{subfigure}
\begin{subfigure}{0.32\textwidth}
    \includegraphics[width=\textwidth]{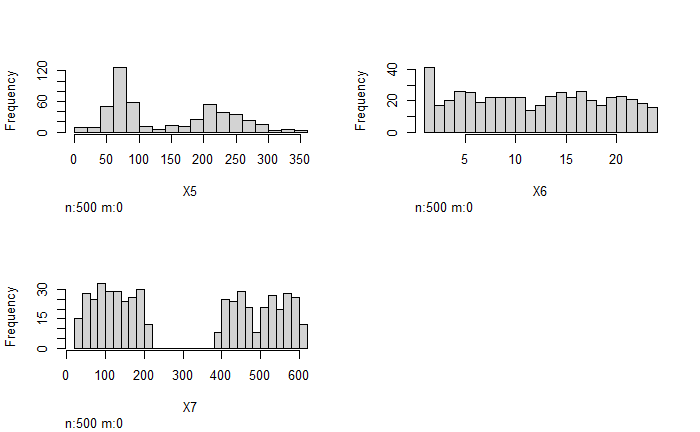}
    \caption{Histograms of the covariates $x$ in the original dataset}
\end{subfigure}
\begin{subfigure}{0.32\textwidth}
    \includegraphics[width=\textwidth]{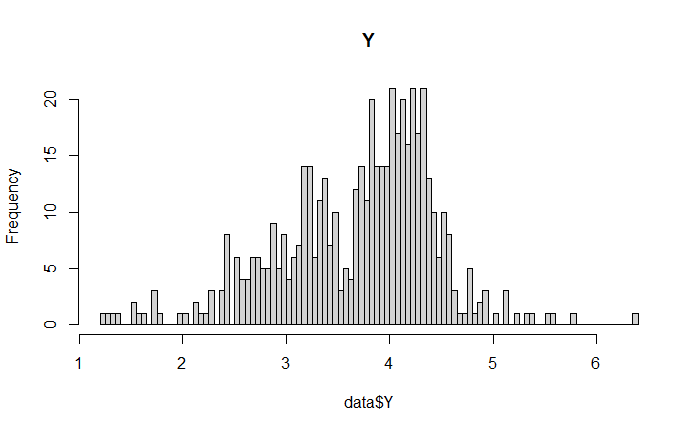}
    \caption{Histogram of the target variable $Y$ in the original dataset}
\end{subfigure}
\caption{Histograms of $X$ and $y$ for the NO2 dataset}
\end{figure}

\begin{table}[H]
\caption{Descriptive statistics of the dataset} 
\scriptsize
\begin{center}
\begin{tabular}{ | c|| c| c || c || c || c || c | } 
Variable & Min. & 1st Qu. & Median & Mean & 3rd Qu. & Max. \\
\hline
Y & 1.22 & 3.21 & 3.85 & 3.7 & 4.22 & 6.4\\
X1 & 4.13 & 6.18 & 7.43 & 6.97 & 7.79 & 8.35\\
X2 & -18.6 & -3.9 & 1.1 & 0.85 & 4.9 & 21.1\\
X3 & 0.3 & 1.67 & 2.8 & 3.06 & 4.2 & 9.9\\
X4 & -5.4 & -0.2 & 0 & 0.15 & 0.6 & 4.3\\
X5 & 2 & 72 & 97 & 143.37 & 220 & 359\\
X6 & 1 & 6 & 12.5 & 12.38 & 18 & 24\\
X7 & 32 & 118.75 & 212 & 310.47 & 513 & 608\\

\hline
\end{tabular}
\label{table8}
\end{center}
\end{table}

\subsubsection{Predictive performances metrics}

The following Figures show the predictive performance metrics for the 5 datasets. We can see that the previous results on the illustrative application are confirmed: GOLIATH outperforms the results.   

\paragraph{Abalone dataset}
The following Figures show the predictive performance metrics for the Abalone dataset.
\begin{figure}[H]
\centering
\begin{subfigure}{0.32\textwidth}
    \includegraphics[width=\textwidth]{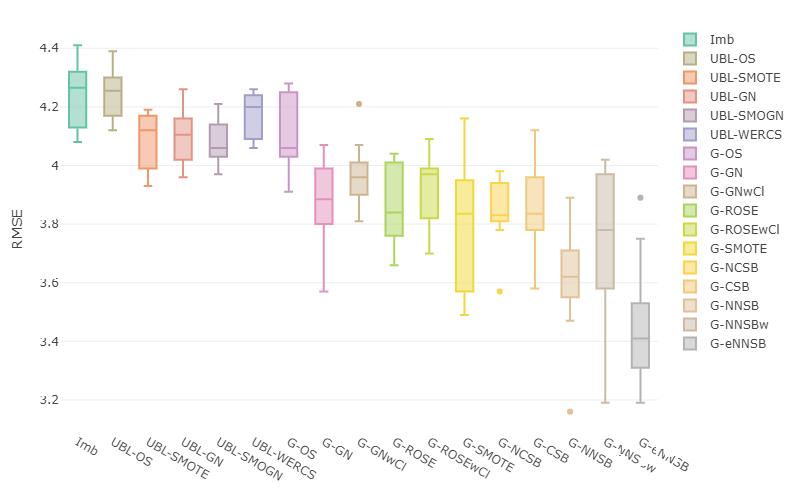}
    \caption{RMSE Boxplots}
\end{subfigure}
\begin{subfigure}{0.32\textwidth}
    \includegraphics[width=\textwidth]{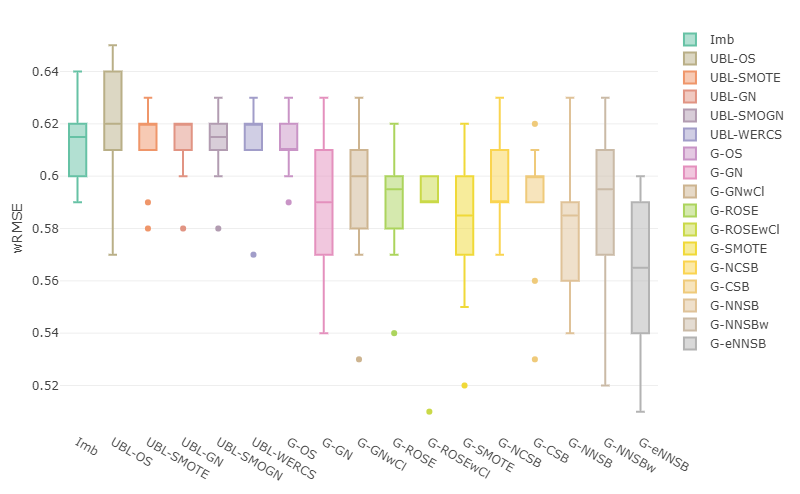}
    \caption{weighted-RMSE bowplots}
\end{subfigure}
\begin{subfigure}{0.32\textwidth}
    \includegraphics[width=\textwidth]{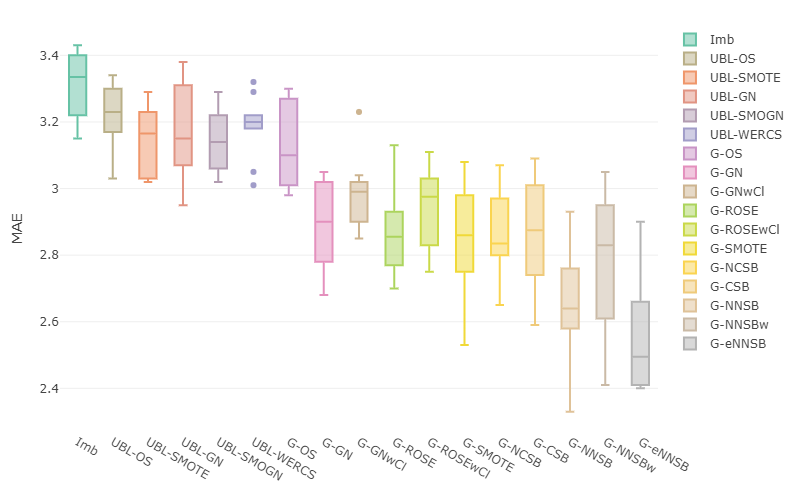}
    \caption{MAE Boxplots}
\end{subfigure}
\caption{Boxplots of predictive performance metrics for Abalone Dataset}
\end{figure}

\paragraph{Bank8FM dataset}
The following Figures show the predictive performance metrics for the Bank8FM dataset.
\begin{figure}[H]
\centering
\begin{subfigure}{0.32\textwidth}
    \includegraphics[width=\textwidth]{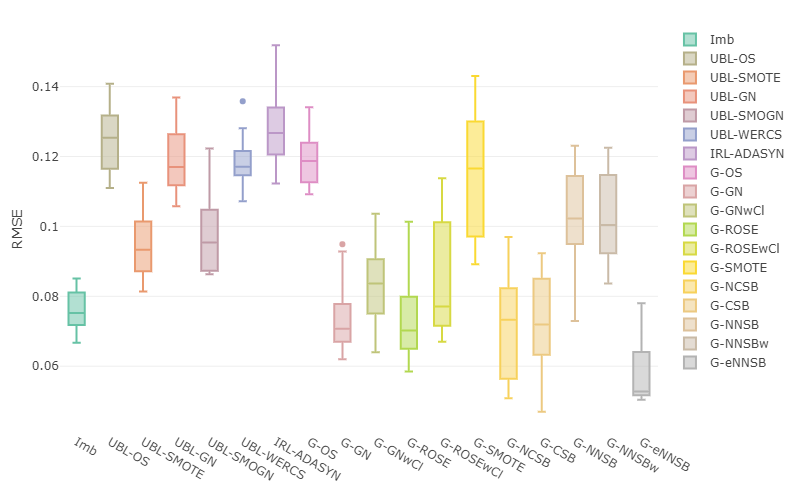}
    \caption{RMSE Boxplots}
\end{subfigure}
\begin{subfigure}{0.32\textwidth}
    \includegraphics[width=\textwidth]{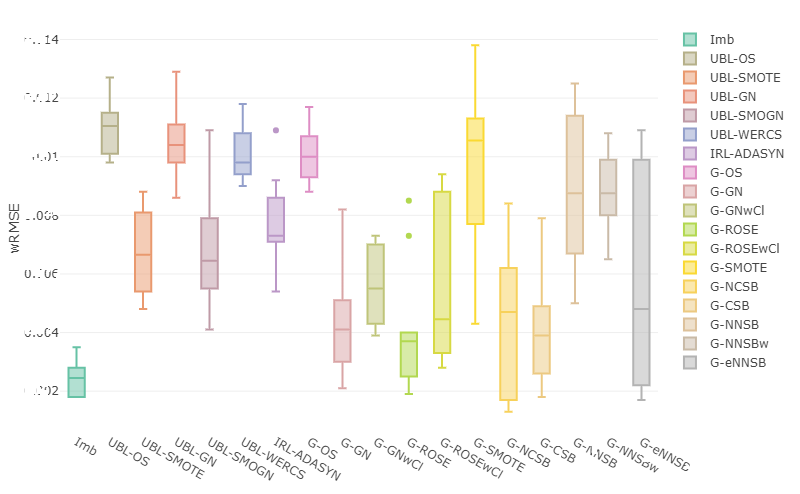}
    \caption{weighted-RMSE bowplots}
\end{subfigure}
\begin{subfigure}{0.32\textwidth}
    \includegraphics[width=\textwidth]{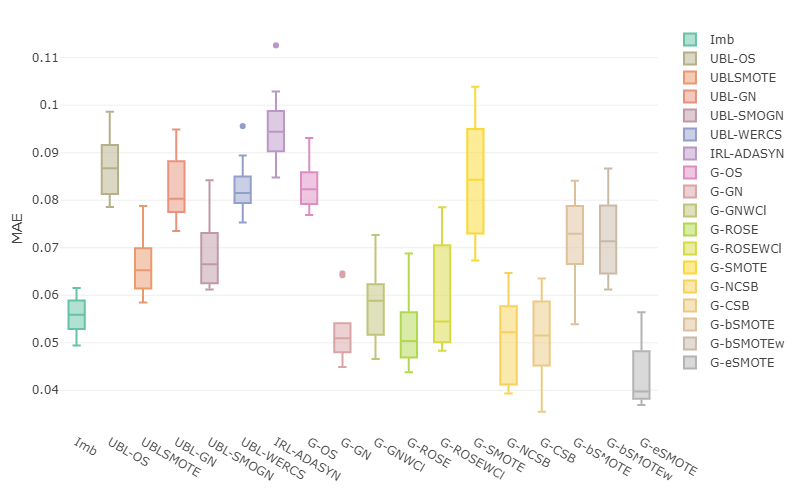}
    \caption{MAE Boxplots}
\end{subfigure}
\caption{Boxplots of predictive performance metrics for Bank8FM Dataset}
\end{figure}

\paragraph{Boston dataset}
The following Figures show the predictive performance metrics for the Boston dataset.
\begin{figure}[H]
\centering
\begin{subfigure}{0.32\textwidth}
    \includegraphics[width=\textwidth]{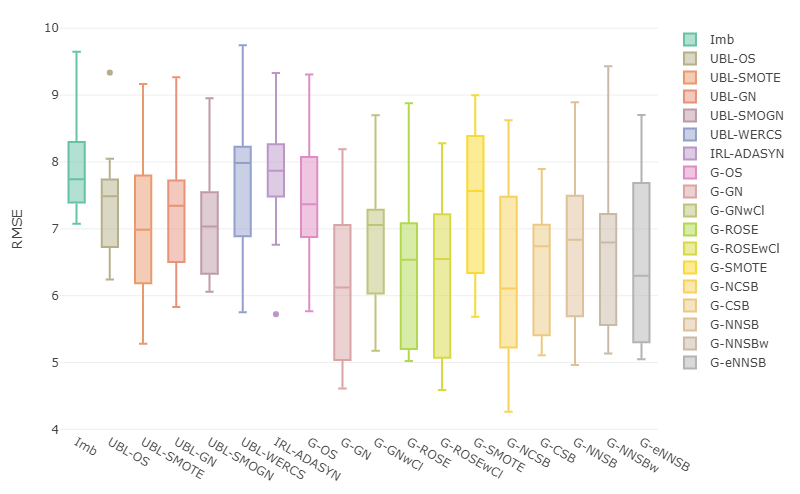}
    \caption{RMSE Boxplots}
\end{subfigure}
\begin{subfigure}{0.32\textwidth}
    \includegraphics[width=\textwidth]{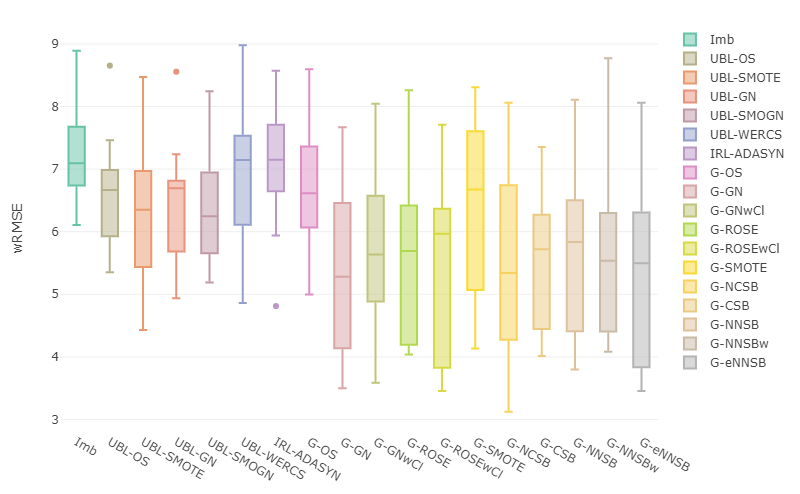}
    \caption{weighted-RMSE bowplots}
\end{subfigure}
\begin{subfigure}{0.32\textwidth}
    \includegraphics[width=\textwidth]{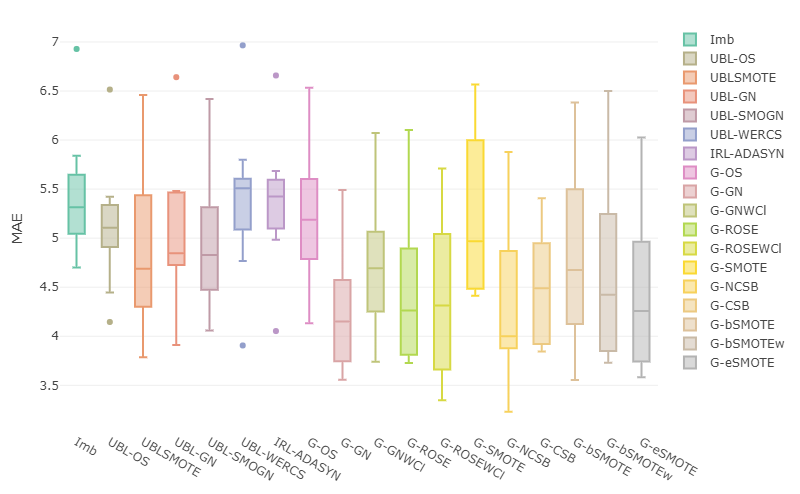}
    \caption{MAE Boxplots}
\end{subfigure}
\caption{Boxplots of predictive performance metrics for Boston Dataset}
\end{figure}

\paragraph{CpuSm dataset}
The following Figures show the predictive performance metrics for the CpuSm dataset.
\begin{figure}[H]
\centering
\begin{subfigure}{0.32\textwidth}
    \includegraphics[width=\textwidth]{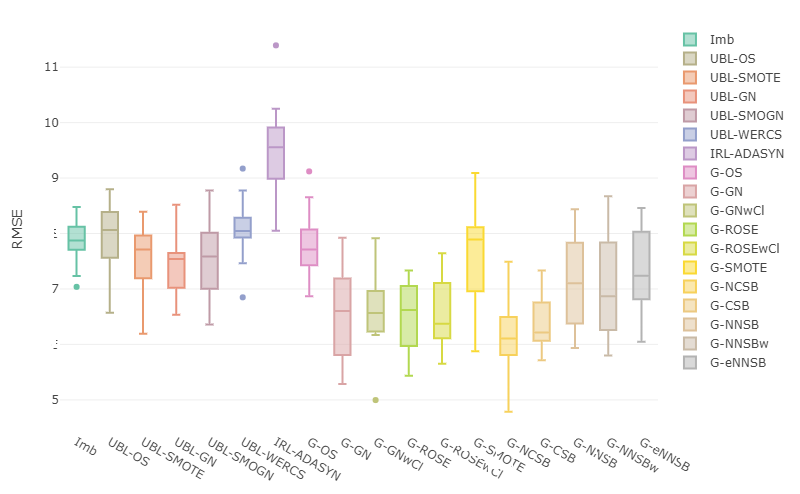}
    \caption{RMSE Boxplots}
\end{subfigure}
\begin{subfigure}{0.32\textwidth}
    \includegraphics[width=\textwidth]{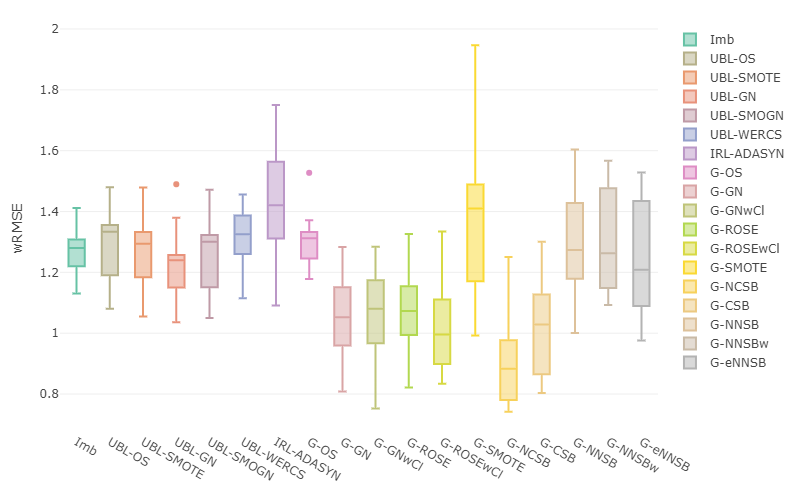}
    \caption{weighted-RMSE bowplots}
\end{subfigure}
\begin{subfigure}{0.32\textwidth}
    \includegraphics[width=\textwidth]{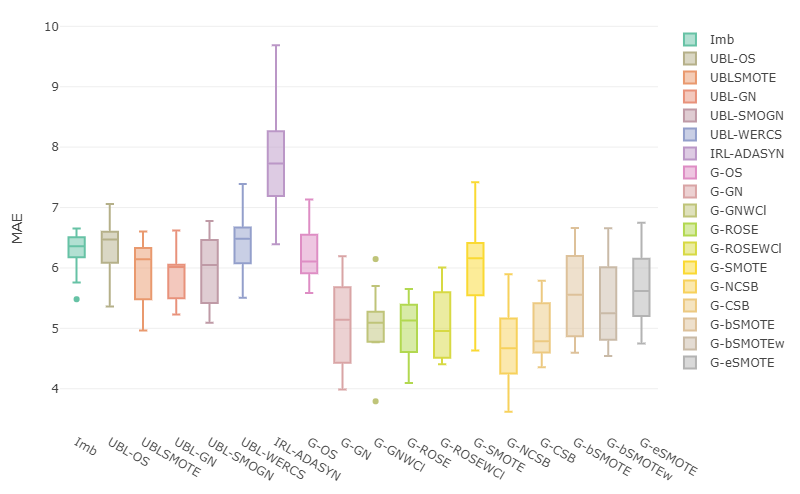}
    \caption{MAE Boxplots}
\end{subfigure}
\caption{Boxplots of predictive performance metrics for CpuSm Dataset}
\end{figure}

\paragraph{NO2 dataset}
The following Figures show the predictive performance metrics for the NO2 dataset.
\begin{figure}[H]
\centering
\begin{subfigure}{0.32\textwidth}
    \includegraphics[width=\textwidth]{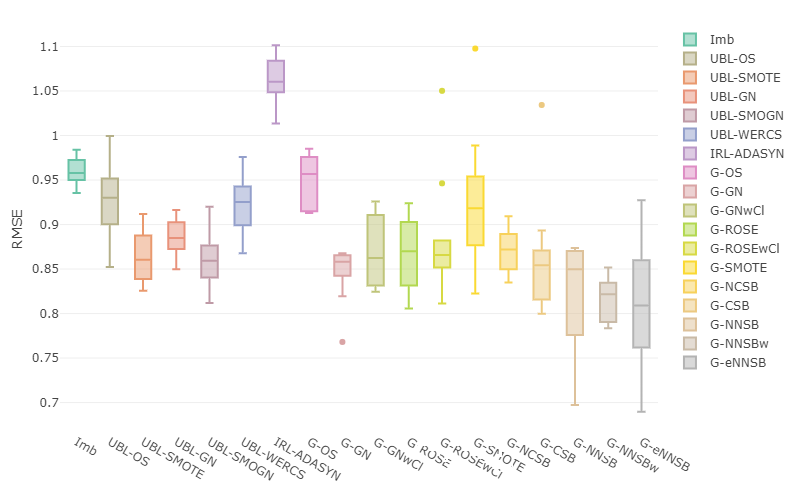}
    \caption{RMSE Boxplots}
\end{subfigure}
\begin{subfigure}{0.32\textwidth}
    \includegraphics[width=\textwidth]{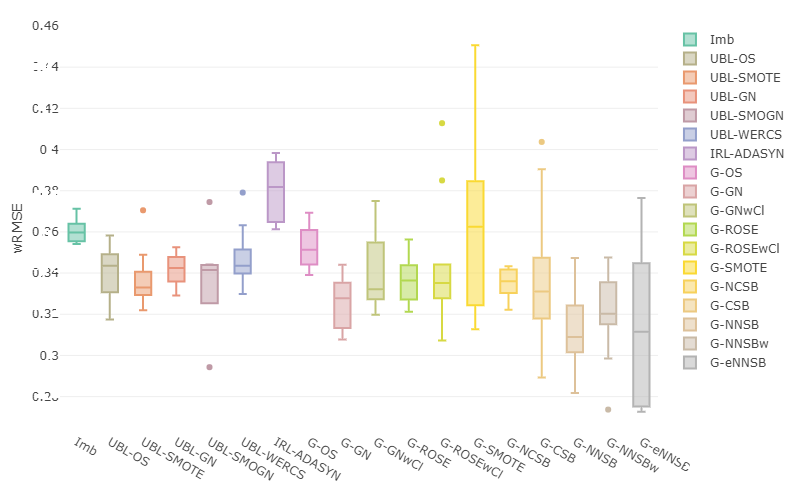}
    \caption{weighted-RMSE bowplots}
\end{subfigure}
\begin{subfigure}{0.32\textwidth}
    \includegraphics[width=\textwidth]{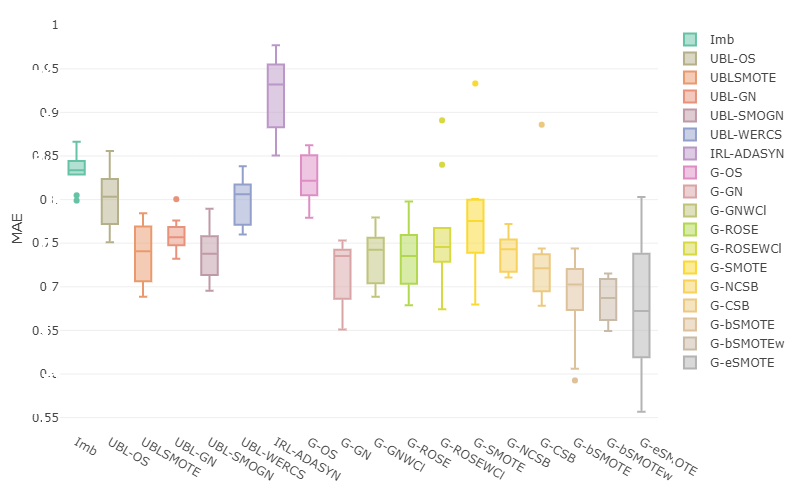}
    \caption{MAE Boxplots}
\end{subfigure}
\caption{Boxplots of predictive performance metrics for NO2 Dataset}
\end{figure}

\newpage

\bibliography{references}


\end{document}